\documentclass{article}

\PassOptionsToPackage{numbers,sort&compress}{natbib}
\usepackage[main,final]{neurips_2026}

\usepackage[utf8]{inputenc}
\usepackage[T1]{fontenc}
\pdfmapfile{+fonts/cm-super-tt.map}
\usepackage{hyperref}
\usepackage{url}
\usepackage{booktabs}
\usepackage{amsfonts}
\usepackage{amsmath}
\usepackage{amssymb}
\usepackage{nicefrac}
\usepackage{microtype}
\usepackage{xcolor}
\usepackage{graphicx}
\usepackage{algorithm}
\usepackage{algpseudocode}
\usepackage{multirow}
\usepackage{pifont}
\usepackage{tikz}
\usepackage{pgfplots}
\usepackage{xspace}
\usepackage{tabularx}
\usepackage{wrapfig}
\usepackage{capt-of}
\usepackage[capitalize,nameinlink]{cleveref}
\usepackage{etoolbox}

\makeatletter
\patchcmd{\NAT@citexnum}{--\NAT@penalty}{-\NAT@penalty}{}%
  {\PackageError{procompnav}{Unable to set citation range hyphen}{}}
\makeatother

\definecolor{urlblue}{RGB}{0,85,180}
\hypersetup{
  colorlinks=true,
  hypertexnames=false,
  linkcolor=urlblue,
  citecolor=urlblue,
  urlcolor=urlblue,
  pdfauthor={Junhyuk Kwon, Seungjoon Lee, Hyejin Park, Kyle Min, Jungseul Ok},
  pdftitle={ProCompNav: Proactive Instance Navigation with Comparative Judgment for Ambiguous User Queries},
  pdfsubject={NeurIPS 2026 Main Track camera-ready paper},
  pdfkeywords={Embodied AI, Instance Navigation, Vision-Language Navigation, Human-Robot Interaction, Large Language Models}
}

\newcommand{\cmark}{\textcolor{green!60!black}{\ding{51}}}
\newcommand{\xmark}{\textcolor{red}{\ding{55}}}
\pgfplotsset{compat=1.18}

\definecolor{lightgray}{gray}{0.93}
\definecolor{promptred}{RGB}{220,50,47}
\newcommand{\promptvar}[1]{\textcolor{promptred}{\{#1\}}}
\newcommand{\promptbox}[1]{%
  \noindent
  \colorbox{lightgray}{%
    \parbox{0.97\linewidth}{%
      \ttfamily\small
      #1
    }%
  }%
}

\newcommand{\OursName}{ProCompNav\xspace}
\newcommand{\rebuttalrev}[1]{#1}
\newcommand{\iclrrev}[1]{#1}
\newcommand{\sj}[1]{#1}

\title{ProCompNav: Proactive Instance Navigation with Comparative Judgment for Ambiguous User Queries}

\author{
  \textbf{Junhyuk Kwon\textsuperscript{1}},
  \textbf{Seungjoon Lee\textsuperscript{2}},
  \textbf{Hyejin Park\textsuperscript{1}},
  \textbf{Kyle Min\textsuperscript{3}},
  \textbf{Jungseul Ok\textsuperscript{1,2}\thanks{Corresponding author.}}
  \\
  \textsuperscript{1}GSAI, POSTECH\ \ \ \ \textsuperscript{2}CSE, POSTECH\ \ \ \ \textsuperscript{3}Oracle
  \\
  \texttt{\{treejhk, sjlee1218, parkebbi2, jungseul\}@postech.ac.kr}
  \\
  \texttt{kyle.min@oracle.com}
  \\
  \href{https://tree-jhk.github.io/procompnav/}
  {\texttt{https://tree-jhk.github.io/procompnav/}}
}

\begin{document}

\setcounter{footnote}{1}
\maketitle
\setcounter{footnote}{0}

\begin{abstract}
In human-robot interaction, agentic robots should resolve ambiguous user requests through proactive clarification while keeping user burden low.
We study this challenge in interactive instance navigation, where an agent must identify and reach the user-intended target among similar objects.
Existing approaches ask open-ended questions and select a candidate that strongly matches the accumulated user-provided information.
However, this process can require lengthy user responses and lead the agent to make premature decisions without distinguishing the target from distractors.
To address these limitations, we propose Proactive Instance Navigation with Comparative Judgment (ProCompNav), a two-stage framework that first constructs a candidate pool and then identifies the target through Recursive Comparative Judgment (RCJ). 
RCJ iteratively narrows the pool by selecting an attribute-value pair that divides the candidates, asking the user a binary question, and removing inconsistent candidates, without requiring an attribute unique to the target.
ProCompNav substantially improves Success Rate and reduces Response Length on an interactive navigation benchmark, and achieves the highest Success Rate among evaluated methods on a non-interactive benchmark.
In two human studies, participants preferred ProCompNav's binary questions to open-ended questions and its batched delivery to interleaved interaction.
Our results highlight the need to jointly consider task performance and user-friendly interaction strategies when designing agentic robots.
\end{abstract}

\section{Introduction}\label{sec:intro}


LLM-based agents support human-agent collaboration on digital tasks by clarifying user intent when requests are incomplete or ambiguous~\citep{wang2025learning,wu2025collabllm,shao2026collaborative}.
This challenge also arises in agentic robotics, where human requests may leave task goals or preferences unspecified, requiring robots to ask follow-up questions as they plan and act~\citep{huang2023inner,ren2023robots}.
In human-robot interaction, clarification should help robots accomplish the intended task while limiting user effort.
For example, ``Find the cabinet'' gives a robot an object category to search for, but several cabinets may fit the request even when the user has a particular one in mind.
What needs to be clarified can then depend on the objects the robot encounters.
We focus on interactive instance navigation~\citep{taioli2025collaborative}, aiming to identify and reach the intended instance through proactive clarification while keeping user burden low.


\raggedbottom
\begin{figure}[!t]
\vspace{-0.5em}

\noindent
\begin{minipage}{\textwidth}
\centering
\small
\setlength{\tabcolsep}{4.1pt}
\renewcommand{\arraystretch}{1.0}

\captionof{table}{Comparison of three disambiguation strategies on CoIN-Bench. We report Success Rate (SR), average total Response Length (RL) per episode, and average Number of Questions (NQ) over episodes with user interaction.}
\label{tab:user_burden}

\vspace{0.25em}

\resizebox{1.0\textwidth}{!}{%
\begin{tabular}{lccccccccccc}
\toprule
\multirow{2}{*}{Decision Strategy} &
\multicolumn{2}{c}{Stage} &
\multicolumn{3}{c}{Val Seen} &
\multicolumn{3}{c}{Val Seen Synonyms} &
\multicolumn{3}{c}{Val Unseen} \\
\cmidrule(lr){2-3}\cmidrule(lr){4-6}\cmidrule(lr){7-9}\cmidrule(lr){10-12}
& Pool & Compare
& SR $\uparrow$ & RL $\downarrow$ & NQ $\downarrow$
& SR $\uparrow$ & RL $\downarrow$ & NQ $\downarrow$
& SR $\uparrow$ & RL $\downarrow$ & NQ $\downarrow$ \\
\midrule
(a) Independent Matching~\citep{taioli2025collaborative}
& \xmark & \xmark
& 10.5 & 109.5 & 1.2
& 15.3 & 129.2 & 1.2
& 8.9 & 122.8 & 1.3 \\
(b) Pooled Independent Matching
& \cmark & \xmark
& 17.5 & 460.2 & 4.1
& 22.0 & 519.2 & 4.3
& 13.3 & 467.7 & 4.0 \\
\textbf{(c) Comparative Judgment (Ours)}
& \cmark & \cmark
& \textbf{23.7} & \textbf{4.2} & 2.2
& \textbf{28.1} & \textbf{4.3} & 2.2
& \textbf{17.0} & \textbf{4.2} & 2.3 \\
\bottomrule
\end{tabular}%
}
\end{minipage}

\vspace{0.15em}

\noindent
\begin{minipage}{\textwidth}
\centering
\includegraphics[width=1.0\linewidth,trim=1 0 1 0,clip]{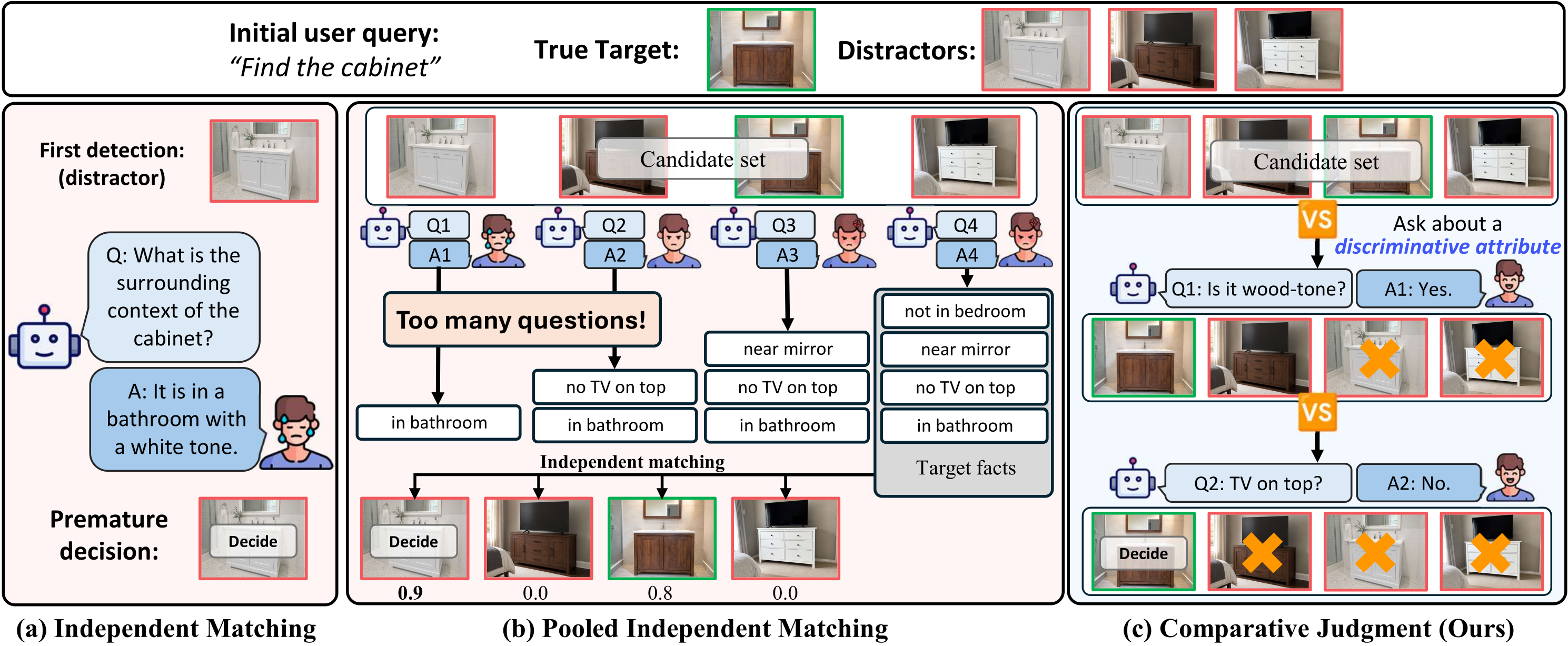}

\vspace{-0.2em}

\captionof{figure}{
Three strategies for instance navigation under an ambiguous user query. (a) \textit{Independent Matching}~\citep{taioli2025collaborative} evaluates each candidate independently. (b) \textit{Pooled Independent Matching} collects multiple candidates, accumulates user responses about each candidate, and applies independent matching. (c) \textit{Comparative Judgment} (Ours) constructs a candidate pool and asks discriminative binary questions based on candidate contrasts.}
\label{fig:comparative_vs_absolute}
\end{minipage}
\end{figure}

One approach to identifying the intended instance is \textit{independent matching}, in which the agent asks the user open-ended questions about attributes of each detected candidate (e.g., color) and scores candidates based on the accumulated responses (e.g., blue)~\citep{taioli2025collaborative}. However, this can lead to premature selection of a distractor sharing many target attributes~(\Cref{fig:comparative_vs_absolute}(a)). A natural way to mitigate this is \textit{pooled independent matching}, which collects multiple candidates before making a decision and applies independent matching to each candidate. By accumulating more information, it increases the chance of distinguishing the target from distractors. However, the accumulated information may still be insufficient to distinguish similar distractors, while asking about every candidate substantially increases both the number of questions and the length of user responses~(\Cref{fig:comparative_vs_absolute}(b)).

\sj{
We propose Proactive Instance Navigation with Comparative Judgment (ProCompNav), an interactive instance navigation framework for identifying the user-intended target with robustness to distractors and low user burden.
ProCompNav first constructs a candidate pool, and then disambiguates the target from the pool through Recursive Comparative Judgment (RCJ) (Figure~\ref{fig:comparative_vs_absolute}(c)). RCJ iteratively narrows the pool by (i) selecting an attribute (e.g., material) and its value (e.g., metal) that divide the remaining candidates into two groups, (ii) asking the user whether the target has that value, and then (iii) removing candidates inconsistent with the answer.
ProCompNav improves robustness to distractors by asking questions on attribute-value pairs that distinguish candidates even when they share many other attributes.
ProCompNav also reduces user burden in two ways. First, each discriminative question can eliminate multiple candidates with a single answer, reducing the number of questions needed. Second, \OursName{} asks its questions consecutively after collecting candidates, reducing repeated, disruptive interruptions that draw users' attention back to the agent~\citep{fitz2019batching,bailey2006attention}.
}


\sj{
We summarize our contributions as follows:
(1) We propose ProCompNav, an interactive instance navigation framework that identifies the user-intended target despite distractors while reducing user burden. We achieve this by our core component, RCJ, which compares candidates, makes discriminative binary questions, and removes candidates inconsistent with user answers.
(2) We evaluate ProCompNav on navigation benchmarks and in two human studies.
On interactive CoIN-Bench, ProCompNav improves success rate while reducing response length.
It also achieves the highest success rate among evaluated methods on the non-interactive TextNav benchmark.
In our human studies, participants preferred ProCompNav's binary questions to open-ended questions and its batched delivery to interleaved interaction~(\Cref{app:user_study}).
ProCompNav shows how agentic robots can use their own observations to guide dialogue, helping people clarify their intent with brief answers.
}

\newpage
\flushbottom

\section{Related Works}
\subsection{Language-driven instance navigation}

Unlike the widely studied object navigation task~\citep{gadre2023cows,zhou2023esc,yu2023l3mvn,cai2024bridging,yokoyama2024vlfm,wu2024voronav,zhang2024trihelper,huang2024gamap,yin2024sg,long2024instructnav,yin2025unigoal,zhang2025multi,zhang2025function,cao2025cognav,zhou2025beliefmapnav,yang2026efficientnav,wang2026trajrag}, where reaching any instance of the target category is sufficient, language-driven instance navigation requires identifying a specific target instance among same-category distractors in a 3D environment. Training-based methods learn navigation policies from observations and descriptions~\citep{sun2024prioritized}, but often require substantial training compute and generalize poorly to unseen environments. Training-free methods have therefore emerged as alternatives~\citep{jang2026context,yang20253d,yin2025unigoal,ziliotto2025tango}. Most existing methods match observed candidates to a fixed target description rather than explicitly comparing the current candidates, leaving them vulnerable to distractors with similar attributes~\citep{jang2026context,yin2025unigoal,ziliotto2025tango}. \iclrrev{Specifically, UniGoal~\citep{yin2025unigoal} matches goal and scene graphs using predefined node, edge, and topology scores, while Context-Nav~\citep{jang2026context} verifies candidates using a predefined set of spatial predicates. Such fixed matching and verification rules do not adapt to the discriminative attributes that distinguish the current candidates.} 3D-Mem~\citep{yang20253d} incrementally builds a visual memory of explored regions and uses a VLM to guide navigation and select the target, but does not explicitly extract discriminative attributes from candidate contrasts, potentially leaving the target ambiguous among similar distractors. Moreover, most existing methods assume a detailed target description is provided upfront, limiting their applicability to ambiguous queries that require user clarification.

\subsection{\iclrrev{Interactive target disambiguation}}
As user requests are often ambiguous, a growing body of work studies how embodied agents can resolve task ambiguity through user clarification~\citep{lin2025ask,ren2023robots}. However, existing object-disambiguation methods address candidates visible in a scene or found within a single room, rather than searching an unknown environment to discover them~\citep{pramanick2022doro,yang2022interactive}. Similarly, AmbResVLM~\citep{chisari2025robotic} resolves ambiguous task descriptions through user clarification in a visible scene. Unlike these methods, AIUTA~\citep{taioli2025collaborative} explores unknown environments while disambiguating the target through user dialogue. However, AIUTA independently evaluates each encountered candidate against accumulated user responses, potentially mistaking a distractor sharing many target attributes for the true target. Its open-ended questions also require users to provide descriptive responses. In contrast, ProCompNav collects candidates before disambiguation and selects discriminative attributes from candidate contrasts, enabling binary questions that distinguish similar distractors with short user responses.

\section{Problem Formulation}\label{sec:problem}

We study an interactive instance navigation task in an unknown 3D environment, where a robot starts from an arbitrary location and must identify the user-intended target instance $T^*$. The robot navigates using RGB-D observations with a restricted $30^\circ$ field of view (FOV).
At the beginning of each episode, the robot receives minimal natural language input: an open-vocabulary category-$c$ (e.g., ``cabinet''). 
Let $\mathcal{O}_c$ denote the set of all instances of category-$c$ in the environment (unknown to the robot). The target is unique, $T^* \in \mathcal{O}_c$, and the remaining instances are distractors $\mathcal{D} = \mathcal{O}_c \setminus \{T^*\}$. 
Since the category-$c$ may not uniquely specify an instance, $T^*$ can be indistinguishable from the distractors $\mathcal{D}$ with minimal information.
To resolve such ambiguities, the robot can engage in a natural language dialogue with the user during navigation, without exchanging any visual information~\citep{taioli2025collaborative}.
At each timestep, the robot can take four discrete actions: \texttt{move\_forward}, \texttt{turn\_left}, \texttt{turn\_right}, and \texttt{stop}, and may optionally ask a free-form natural-language question.
An episode terminates when the robot executes \texttt{stop} or the predefined episode horizon is reached.
Success is declared if the robot executes \texttt{stop} within distance $r$ of $T^*$, within the horizon.

Because distractors $\mathcal{D}$ and $T^*$ share many attribute-value pairs, (pooled) independent matching~\citep{taioli2025collaborative} is reliable only when the available target information suffices to distinguish $T^*$ from $\mathcal{D}$.
We instead identify $T^*$ through \textit{comparative judgment}: the robot accumulates a candidate pool of category-$c$ instances and, at each round, prunes it with an attribute-value pair that splits the pool into two non-empty groups, until only $T^*$ remains.


\section{Proposed Method}\label{sec:method}

\subsection{Overview}
We propose Proactive Instance Navigation with Comparative Judgment (\textbf{\OursName}), which first constructs a candidate pool and then compares candidates to ask discriminative questions that prune the pool.
In the Pool Construction Stage~(Sec~\ref{sec:stage1}), \OursName explores the environment to construct a candidate pool of category-$c$ instances.
In the Recursive Comparison Stage~(Sec~\ref{sec:stage2}), \OursName replaces independent matching with comparative judgment across candidates, and minimizes user burden by asking only binary yes/no questions that can eliminate multiple candidates.

\begin{figure}[t]
  \centering
  \includegraphics[width=1\linewidth,trim=0 0 0 0,clip]{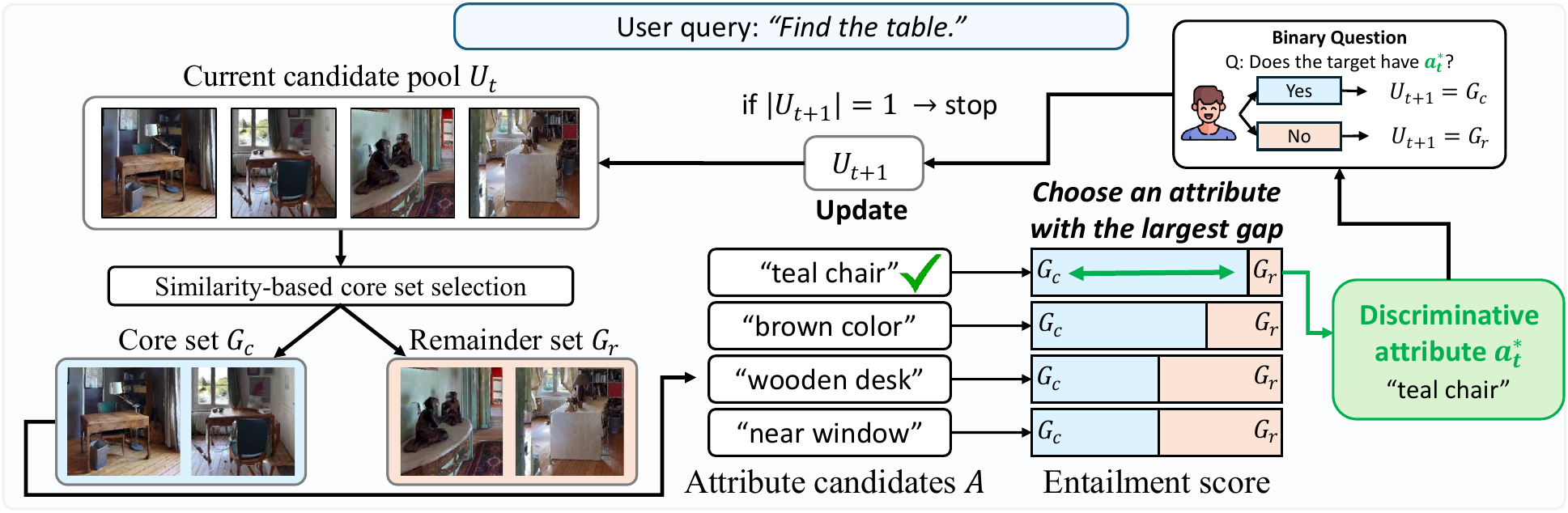}
\vspace{-1.5em}
  \caption{\textbf{Recursive Comparative Judgment.} At iteration $t$, \OursName splits the candidate pool $U_t$ into a core set $G_c$ and a remainder set $G_r$ by similarity. It identifies a discriminative attribute $a_t^*$, that is common in $G_c$ but not in $G_r$. Finally, it asks whether the target has $a_t^*$, and prunes the pool to obtain the next candidate pool $U_{t+1}$ based on the user’s response.}
  \label{fig:stage2}
\end{figure}

\subsection{Pool construction stage}
\label{sec:stage1}
The Pool Construction Stage explores the unknown environment to collect the candidate pool $\mathcal{C}=\{c_i\}_{i=1}^{N}$ of $N$ distinct candidates on which the Recursive Comparison Stage operates. Each candidate $c_i=(I_i, d_i)$ is represented by a multi-view collage image $I_i$ and a multi-view description $d_i$.
For each candidate $i$, the agent maintains an accumulated 3D point cloud $\mathcal{P}_i$ of the instance and a set of RGB views $\mathcal{V}_i$ captured from different viewpoints during exploration.
For each new detection of category-$c$, we extract a 3D point cloud from the detected region. If it sufficiently overlaps with the accumulated point cloud $\mathcal{P}_i$ of an existing candidate $i$, the detection is assigned to candidate $i$ and its RGB view is added to $\mathcal{V}_i$. Otherwise, a new candidate is initialized with its point cloud and view set seeded from this detection.
From $\mathcal{V}_i$, we cluster the views in an embedding space into $K$ clusters and select a representative per cluster, arrange the $K$ representatives into the multi-view collage $I_i$, and prompt an MLLM to produce the multi-view description $d_i$. The description summarizes the candidate's attribute-value pairs---attributes (e.g., color, nearby objects) and their values (e.g., blue, next to a TV)---aggregated across the $K$ views, capturing pairs that may be missed from any single viewpoint.
Once $|\mathcal{C}| \geq N_{\min}$, the Pool Construction Stage terminates and the agent transitions to the Recursive Comparison Stage.
Implementation details and example multi-view candidates are provided in Appendix~\ref{app:pool_construction_details}.

\subsection{Recursive comparison stage}\label{sec:stage2}
We propose \textbf{Recursive Comparative Judgment} (RCJ, \cref{fig:stage2}), which identifies the target $T^*$ by iteratively pruning the candidate pool $\mathcal{C}$ through binary questions on attribute-value pairs that contrast candidates.
At each interaction round, \OursName extracts an attribute-value pair---a \textit{Discriminative Attribute (DA)} that one group of candidates in the current pool has but the rest lack---asks the user whether $T^*$ has it, and removes the inconsistent group.
Crucially, RCJ does not require an attribute that uniquely identifies $T^*$ at any round.
Each round only needs an attribute-value pair that splits the current pool into two non-empty groups.
Formally, let $U_t \subseteq \mathcal{C}$ denote the active candidate set at interaction round $t$, initialized as $U_0=\mathcal{C}$. At iteration $t$, \OursName discovers a DA $a_t^*$ that partitions $U_t$. After asking the user whether $T^*$ possesses $a_t^*$ and receiving their binary answer, \OursName updates the active candidate set to $U_{t+1}$ by retaining only the candidates consistent with this answer. This recursive pruning continues until $|U_t|=1$. We next describe how core selection helps find an attribute that partitions $U_t$.

\subsubsection{Similarity-based core set selection}\label{subsubsec:core_set}
\iclrrev{When $|U_t|>2$, we select a similar pair as the core set $G_c$ and place the other candidates in the remainder set $G_r = U_t \setminus G_c$.}
\iclrrev{In our ablation, this core-remainder split yields higher SR than KMeans grouping (Table~\ref{tab:pipeline_ablation}, Appendix~\ref{app:core_set_ablation}).}
\iclrrev{We measure the cohesion of a subset $V \subseteq U_t$ by its mean pairwise similarity:}
\begin{align*}
    \rho(V) = \frac{1}{|V|(|V|-1)} \sum_{\substack{i,j \in V, i \neq j}} S(i,j) \;,
\end{align*}
where $S(i,j)$ is the pairwise similarity between instances $i$ and $j$ (defined below).

\iclrrev{We use greedy peeling: starting from $V_0=U_t$, we repeatedly remove the candidate with the lowest total similarity to the others until two candidates remain:}
\iclrrev{\begin{align*}
    V_{l+1} = V_l \setminus \{i^\prime\}, \quad \text{where } i^\prime = \arg\min_{i \in V_l} \sum_{\substack{j \in V_l, j\neq i}} S(i,j) \;.
\end{align*}}
\iclrrev{We take the surviving pair as the core set, $G_c=V_{|U_t|-2}$, and set $G_r=U_t\setminus G_c$. Removing the least similar candidate makes $\rho$ non-decreasing along the peeling path, so the surviving pair has the highest $\rho$ on that path.}
\iclrrev{When $|U_t|=2$, we skip peeling and treat each candidate as a singleton group. We select the attribute with the largest entailment-score difference between them, and the binary answer updates the active set as described below.}

We compute the pairwise similarity $S(i,j)$ by averaging text and visual similarities:
\begin{align*}
    S(i,j) = \tfrac{1}{2}\left(\left\langle \hat{e}^{\text{text}}_{i}, \hat{e}^{\text{text}}_{j} \right\rangle + \left\langle \hat{e}^{\text{img}}_{i}, \hat{e}^{\text{img}}_{j} \right\rangle\right) \;,
\end{align*}
where $\hat{e}^{\text{text}}$ and $\hat{e}^{\text{img}}$ denote the $\ell_2$-normalized textual and visual embeddings extracted from the instance's description and image, respectively.

\subsubsection{Discriminative attribute (DA) discovery}\label{subsubsec:da_discovery}
To identify the target $T^*$, we aim to discover a DA $a_t^*$ that contrasts the core set $G_c$ against the remainder set $G_r$ and enables effective pruning with a yes/no question.
This requires judging whether each instance description supports a candidate attribute.
We use a Natural Language Inference (NLI) model to verify them: given the description $d_i$ of an instance $i$ and a hypothesis of the form ``instance $i$ has attribute $a$'', the NLI model provides a standardized judgment (entails, contradicts, or neutral) that yields stable scores for quantifying group-level contrast.

Our procedure consists of three steps:
(i) we extract a candidate set of attributes $\mathcal{A}$ from captions in $G_c$ using an LLM; (ii) we score each attribute $a \in \mathcal{A}$ on every instance $i \in G_c \cup G_r$ using the NLI-based entailment score $s(d_i, a)$; (iii) we select the DA $a_t^*$ that maximizes the contrast between $G_c$ and $G_r$:
\begin{align*}
    a_t^* = \arg\max_{a \in \mathcal{A}} \left( \mathbb{E}_{i \in G_c}[s(d_i, a)] - \mathbb{E}_{j \in G_r}[s(d_j, a)] \right) \;.
\end{align*}
The entailment score $s(d_i, a)$ is derived from the NLI logits. Detailed definitions of the scoring function and implementation settings are provided in Appendix~\ref{app:entailment}.

\subsubsection{Attribute-guided group refinement}\label{subsubsec:refinement}
Although $a_t^*$ is selected to maximize the contrast between $G_c$ and $G_r$, the candidate attribute set $\mathcal{A}$ is extracted from $G_c$ only, so $a_t^*$ may also be possessed by candidates in $G_r$.
We therefore refine the remainder set $G_r$ by re-evaluating each candidate $i \in G_r$ via the NLI-based entailment score $s(d_i, a_t^*)$, and moving $i$ to $G_c$ if $s(d_i, a_t^*) \ge \tau$, where $\tau$ is a conservative threshold.
This can add candidates to the core before asking the question.
A partition is valid only when both groups are non-empty. If refinement produces an empty group, we try the next attribute in descending order of entailment gap.
If no attribute yields two non-empty groups, we ask about the last attribute.

\subsubsection{Interactive pruning and re-exploration}\label{subsubsec:pruning}
We then ask the user a binary question of whether the target $T^*$ possesses the selected DA $a_t^*$.
Based on the user's response, we update the active candidate pool as:
\begin{equation}
U_{t+1} =
\begin{cases}
G_c, & \text{if the user answers \textbf{Yes},} \\
G_r, & \text{otherwise (\textbf{No}),}
\end{cases}
\end{equation}
thereby pruning the candidate pool according to the answer.
Once the candidate pool $U_{t+1}$ is finalized, we identify the target if $|U_{t+1}|=1$. Otherwise, if $|U_{t+1}| \ge 2$, we proceed to the next RCJ round. When no attribute yields a valid split, a \textbf{Yes} answer leaves the pool unchanged.

When the user answers \textbf{No} but the remainder set is empty, \textit{i.e.}, $|G_r|=0$, the initial candidate pool may not contain the target.
In this case, we resume the Pool Construction Stage to add one more candidate, then pre-prune the candidate pool with the target's facts collected from previous RCJ rounds before resuming RCJ.

\section{Experiments}\label{sec:exp}
\subsection{Benchmarks and implementation details}
\label{sec:exp_setup}
\paragraph{Benchmarks} To demonstrate that \OursName is applicable to both ambiguous and detailed user queries, we evaluate it on two simulated benchmarks: CoIN-Bench~\citep{taioli2025collaborative} and text-goal navigation (TextNav)~\citep{sun2024prioritized}.
Both benchmarks feature multi-instance scenes but differ in initial user query specificity and the availability of user interaction.
In CoIN-Bench, the agent is given only a coarse category at episode start, making interactive disambiguation necessary to identify the correct target instance; we follow its standard evaluation splits (\textit{Val Seen}, \textit{Val Seen Synonyms}, \textit{Val Unseen}), and simulate user responses with an MLLM that has access to the image of the target instance.
In contrast, in TextNav, the agent is given a detailed textual description of the target instance at episode start, yielding a non-interactive setting.

\paragraph{Evaluation metrics}
We report Success Rate (SR) and Success weighted by Path Length (SPL) as the primary evaluation metrics.
SR measures the proportion of episodes in which the agent successfully reaches the target, while SPL evaluates exploration efficiency by comparing the agent's path length to the optimal path.
In interactive settings, we additionally report Response Length (RL) and Number of Questions (NQ) as user-burden metrics: RL denotes the total response length per episode, measured as the token count of the user-simulator MLLM responses, and NQ averages questions over episodes with interaction to measure question burden when users are involved.

\paragraph{Implementation details}
We use a single Qwen3-VL-8B~\citep{bai2025qwen3} model for both the text-only LLM and MLLM modules, and adopt VLFM~\citep{yokoyama2024vlfm} as the exploration backbone, consistent with AIUTA~\citep{taioli2025collaborative}.
We initiate the Recursive Comparison Stage once the candidate pool size reaches $N_{\min}=5$ on CoIN or $N_{\min}=3$ on TextNav. If it has not started by step 400 on CoIN or step 600 on TextNav, we start it at that step.
We use DeBERTa-v3-large~\citep{he2021debertav3,laurer2024less} as the NLI verifier for scoring attribute entailment in the Recursive Comparison Stage.
We set the maximum episode length to 500 for CoIN following CoIN-Bench and to 1000 for TextNav following prior work.
On CoIN, we use a four-question budget, except for Pooled Independent Matching (PIM), which has no question limit so it can gather information from every candidate as a strong baseline.
An episode is considered successful if the agent executes \texttt{stop} within 1\,m of the target instance.
On CoIN, the Recursive Comparison Stage extracts attribute candidates from the collected candidate set $U_t$; on TextNav, we disable the user-question module and instead extract attributes directly from the detailed initial goal (see Appendix~\ref{app:ours_textnav}).
We further apply loop-escape and occasional 360$^\circ$ rotation heuristics (see Appendix~\ref{app:exploration}).
We report the per-episode computational cost of AIUTA* and \OursName in Appendix~\ref{app:computational_cost}.

\raggedbottom
\subsection{Main Results}
\label{subsubsec:main_results}


\subsubsection{Interactive Instance Navigation}
As shown in Table~\ref{tab:coin_main}, ProCompNav achieves the highest SR across all CoIN-Bench splits, outperforming both training-based methods~\citep{khanna2024goat,sun2024prioritized} and training-free methods that receive detailed target descriptions~\citep{jang2026context,yang20253d}, despite using only category-level input. On Val Seen, ProCompNav achieves a 126\% relative SR improvement over AIUTA$^*$~\citep{taioli2025collaborative}. Table~\ref{tab:user_burden} compares three end-to-end disambiguation strategies. Pooled Independent Matching uses the same pool construction stage as ProCompNav but applies independent matching, improving SR over AIUTA$^*$. This suggests that collecting candidates before making a decision helps avoid premature selection of early-encountered distractors. However, because its questions are generated independently for each candidate, the collected information may not distinguish the target from distractors that share similar attributes. Using the same pool construction stage, ProCompNav further improves SR by selecting attribute-value pairs that split the pool and pruning candidates from binary feedback. ProCompNav also achieves lower RL and NQ than Pooled Independent Matching, consistent with its binary questions and selection of attributes that distinguish the current candidates.

\begin{table}[t]
\centering
\small
\setlength{\tabcolsep}{4pt}
\renewcommand{\arraystretch}{0.9}

\caption{Performance on CoIN-Bench. In the \textit{Judge} column, Indep/Comp indicate whether each method uses independent matching or comparative judgment for target disambiguation. The \textit{Interact} column indicates whether the method interacts with the user for target disambiguation. For fair comparison across models, we include 3D-Mem$^{*}$ and AIUTA$^{*}$, our reproductions using the same MLLM and LLM as \OursName. Results denoted by $^{\dagger}$ are taken from~\citep{taioli2025collaborative} and those denoted by $^{\ddagger}$ are taken from~\citep{jang2026context}.}
\label{tab:coin_main}

\resizebox{\textwidth}{!}{%
\begin{tabular}{lcccccccc}
\toprule
\multirow{2}{*}{Method} &
\multicolumn{2}{c}{Model Condition} &
\multicolumn{2}{c}{Val Seen} &
\multicolumn{2}{c}{Val Seen Synonyms} &
\multicolumn{2}{c}{Val Unseen} \\
\cmidrule(lr){2-3}\cmidrule(lr){4-5}\cmidrule(lr){6-7}\cmidrule(lr){8-9}
& Judge & Interact &
SR $\uparrow$ & SPL $\uparrow$ &
SR $\uparrow$ & SPL $\uparrow$ &
SR $\uparrow$ & SPL $\uparrow$ \\

\midrule
\multicolumn{9}{l}{\textit{\textbf{Training-based Methods}}} \\
Monolithic-GOAT$^{\dagger}$~\citep{khanna2024goat} & - & \xmark
& 6.6 & 3.1
& 13.1 & 6.5
& 0.2 & 0.1 \\
PSL$^{\dagger}$~\citep{sun2024prioritized} & - & \xmark
& 8.8 & 3.3
& 8.9 & 2.8
& 4.6 & 1.4 \\

\midrule
\multicolumn{9}{l}{\textit{\textbf{Training-free Methods (description input)}}} \\
3D-Mem*~\citep{yang20253d} & Comp & \xmark
& 15.9 & 11.0
& 18.9 & 13.6
& 5.7 & 3.9 \\
Context-Nav$^{\ddagger}$~\citep{jang2026context} & Indep & \xmark
& 13.5 & 6.7
& 20.3 & 10.9
& 11.3 & 5.2 \\

\midrule
\multicolumn{9}{l}{\textit{\textbf{Training-free Methods (category input)}}} \\
VLFM$^{\dagger}$~\citep{yokoyama2024vlfm} & Indep & \xmark
& 0.4 & 0.3
& 0.0 & 0.0
& 0.0 & 0.0 \\
AIUTA$^{\dagger}$~\citep{taioli2025collaborative} & Indep & \cmark
& 7.4 & 2.9
& 14.4 & 8.0
& 6.7 & 2.3 \\
AIUTA$^*$ & Indep & \cmark
& 10.5 & 4.5
& 15.3 & 8.4
& 8.9 & 4.0 \\
Pooled Independent Matching & Indep & \cmark
& 17.5 & 5.1
& 22.0 & 8.1
& 13.3 & 5.1 \\
\textbf{\OursName (Ours)} & Comp & \cmark
& \textbf{23.7} & \textbf{7.0}
& \textbf{28.1} & \textbf{8.5}
& \textbf{17.0} & \textbf{6.2} \\
\bottomrule
\end{tabular}%
}

\end{table}


\subsubsection{Non-interactive Instance Navigation}

\begin{table}[H]
\begin{minipage}[t]{0.49\textwidth}
\vspace{0pt}
\iclrrev{ProCompNav achieves the highest SR (28.5\%) among the evaluated TextNav methods, an 8.8\% relative improvement over Context-Nav~\citep{jang2026context} (Table~\ref{tab:textnav_main}). This result shows that candidate-relative comparison remains effective with detailed target descriptions and without predefined goal-to-scene matching rules~\citep{jang2026context,yin2025unigoal}. We evaluate this non-interactive setting on TextNav~\citep{sun2024prioritized}. Instead of asking user questions, ProCompNav extracts attributes from the initial target description. It otherwise retains the collect-and-compare procedure: it constructs a candidate pool, selects goal-derived attributes from candidate contrasts, refines candidate groups, and recursively retains the group supported by each attribute. Thus, user feedback is removed while the core candidate-comparison mechanism remains unchanged (Appendix~\ref{app:ours_textnav}).}
\end{minipage}\hfill
\begin{minipage}[t]{0.49\textwidth}
\vspace{0pt}
\caption{Performance on TextNav. In the \textit{Judgment} column, Indep/Comp indicate whether each method uses independent matching or comparative judgment for target disambiguation. Results denoted by $^{\dagger}$ are taken from~\citep{yin2025unigoal}, those denoted by $^{\ddagger}$ are taken from~\citep{jang2026context}. Methods denoted by $^{*}$ are our reproductions using the same MLLM and LLM as \OursName.}
\label{tab:textnav_main}
\centering
\footnotesize
\setlength{\tabcolsep}{2pt}
\renewcommand{\arraystretch}{1.0}

\resizebox{0.90\linewidth}{!}{%
\begin{tabular}{lccc}
\toprule
Method & Judgment & SR $\uparrow$ & SPL $\uparrow$ \\
\midrule
\multicolumn{4}{l}{\textit{\textbf{Training-based Methods}}} \\
Modular-GOAT$^{\dagger}$~\citep{khanna2024goat} & - & 17.0 & 8.8 \\
PSL$^{\dagger}$~\citep{sun2024prioritized} & - & 16.5 & 7.5 \\
\midrule
\multicolumn{4}{l}{\textit{\textbf{Training-free Methods}}} \\
UniGoal$^{\dagger}$~\citep{yin2025unigoal} & Indep & 20.2 & \textbf{11.4} \\
AIUTA*~\citep{taioli2025collaborative} & Indep & 10.9 & 2.7 \\
3D-Mem*~\citep{yang20253d} & Comp & 14.1 & 9.6 \\
Context-Nav$^{\ddagger}$~\citep{jang2026context} & Indep & 26.2 & 9.1 \\
\textbf{\OursName (Ours)} & Comp & \textbf{28.5} & 6.9 \\
\bottomrule
\end{tabular}%
}

\end{minipage}
\end{table}

\subsubsection{\rebuttalrev{Robustness to User Feedback}}
\begin{table}[H]
\centering\color{black}\small
\caption{Results with changed user feedback on CoIN-Bench Val Seen. $p$ is the probability of IDK or flipped feedback. \iclrrev{Original reports the reference values from Table~\ref{tab:user_burden}.}}
\label{tab:rebuttal_noise}
\resizebox{\textwidth}{!}{%
\begin{tabular}{llccccccccc}
\toprule
& & \multicolumn{3}{c}{Original} & \multicolumn{3}{c}{$p=10\%$} & \multicolumn{3}{c}{$p=20\%$} \\
\cmidrule(lr){3-5}\cmidrule(lr){6-8}\cmidrule(lr){9-11}
Method & Noise & SR $\uparrow$ & RL $\downarrow$ & NQ $\downarrow$ & SR $\uparrow$ & RL $\downarrow$ & NQ $\downarrow$ & SR $\uparrow$ & RL $\downarrow$ & NQ $\downarrow$ \\
\midrule
\multirow{3}{*}{AIUTA$^*$} & Flip & \multirow{3}{*}{10.5} & \multirow{3}{*}{\iclrrev{109.5}} & \multirow{3}{*}{\iclrrev{1.2}} & 11.3 & 99.3 & 1.2 & 11.3 & 107.6 & 1.2 \\
& IDK & & & & 10.2 & 101.8 & 1.4 & 12.4 & 89.7 & 1.6 \\
& Combined & & & & 9.9 & 98.5 & 1.4 & 10.2 & 90.3 & 1.6 \\
\midrule
\multirow{3}{*}{Pooled} & Flip & \multirow{3}{*}{17.5} & \multirow{3}{*}{\iclrrev{460.2}} & \multirow{3}{*}{\iclrrev{4.1}} & 16.6 & 460.9 & 4.2 & 16.9 & 463.5 & 4.2 \\
& IDK & & & & 16.4 & 411.6 & 4.5 & 17.1 & 360.5 & 4.9 \\
& Combined & & & & 17.7 & 421.4 & 4.5 & 16.0 & 374.2 & 5.1 \\
\midrule
\multirow{3}{*}{\textbf{ProCompNav}} & Flip & \multirow{3}{*}{\textbf{23.7}} & \multirow{3}{*}{\textbf{4.2}} & \multirow{3}{*}{2.2} & \textbf{19.9} & \textbf{4.2} & 2.2 & \textbf{17.1} & \textbf{4.1} & 2.2 \\
& IDK & & & & \textbf{20.8} & \textbf{4.7} & 2.5 & \textbf{21.2} & \textbf{4.1} & 2.8 \\
& Combined & & & & \textbf{20.2} & \textbf{4.1} & 2.4 & \textbf{17.3} & \textbf{4.0} & 2.7 \\
\bottomrule
\end{tabular}}
\end{table}

\iclrrev{The MLLM user simulator is not an idealized noise-free user. We model uncertain or incorrect feedback with IDK, Flip, and Combined perturbations at $p\in\{10\%,20\%\}$. IDK returns ``I don't know,'' while Flip reverses ProCompNav's binary feedback and changes only the queried attribute for the open-ended baselines (Appendix~\ref{app:flip_examples}). Combined applies IDK with probability $p$ and otherwise applies Flip with probability $p$. ProCompNav retains the highest SR in every tested condition, including $p=20\%$ (Table~\ref{tab:rebuttal_noise}). IDK mainly increases NQ as RCJ proceeds to the next-ranked attribute. Flip causes a larger SR drop because each binary answer determines which group is pruned, whereas the baselines retain other information in their free-form responses.}

\vspace{-4pt}
\subsection{Effect of Design Choices}

\begin{wraptable}{r}{0.38\textwidth}
\vspace{-12pt}
\centering
\caption{Ablations on CoIN-Bench Val Seen.}
\label{tab:pipeline_ablation}
\vspace{2pt}
\resizebox{\linewidth}{!}{%
\begin{tabular}{lrrrr}
\toprule
Configuration
& SR $\uparrow$ & SPL $\uparrow$
& RL $\downarrow$ & NQ $\downarrow$ \\
\midrule
w/o multi-view aggregation
& 23.1 & 6.3 & 4.2 & 2.1 \\
LLM-only DA (w/o NLI)
& 20.6 & 5.5 & 4.2 & 2.2 \\
w/o similarity-based core set
& \iclrrev{20.6} & \iclrrev{5.8} & \iclrrev{4.5} & \iclrrev{2.4} \\
w/o refinement
& 21.8 & 6.0 & 4.2 & 2.1 \\
\iclrrev{Prompt-only}
& \iclrrev{20.9} & \iclrrev{5.9} & \iclrrev{4.7} & \iclrrev{2.5} \\
\midrule
\textbf{ProCompNav (Ours)}
& \textbf{23.7} & \textbf{7.0}
& 4.2 & 2.2 \\
\bottomrule
\end{tabular}

}
\vspace{-8pt}
\end{wraptable}

Table~\ref{tab:pipeline_ablation} reports the effect of five design choices on the CoIN-Bench Val Seen split.
Disabling multi-view aggregation and using only a single view per candidate drops SR from $23.7$ to $23.1$.
Removing NLI entirely and letting the LLM alone pick a discriminative attribute from the two group descriptions drops SR to $20.6$, showing that adding the entailment LM provides further performance gain on top of the LLM.
\iclrrev{Replacing our core-remainder split with KMeans grouping based on the same averaged text-image similarity reduces SR from 23.7\% to 20.6\%, supporting the effectiveness of our similarity-based core selection.}
Disabling the refinement module drops SR to $21.8$: since the discriminative attribute is selected from the core set $G_c$ only, the remainder group may still contain candidates that match it, and refinement is needed to recover them.
\iclrrev{The prompt-only variant selects an attribute and splits the pool in one LLM call but achieves lower SR. Further question-format and candidate-comparison ablations are in Appendix~\ref{app:binary_ablation}.}

\vspace{-4pt}
\enlargethispage{60pt}
\subsection{\iclrrev{Human Evaluation of Interaction Strategies}}
\vspace{-6pt}

\begin{wraptable}{r}{0.49\textwidth}
\vspace{-14pt}
\centering
\footnotesize
\caption{\iclrrev{Human evaluation of response format (Study~1) and question delivery (Study~2). Response time and Preference denote the median per question and participant counts, respectively.}}
\label{tab:rebuttal_user_study}
{\fontsize{7.5}{8.4}\selectfont
\setlength{\tabcolsep}{0.5pt}
\renewcommand{\arraystretch}{1.0}
\begin{tabular*}{\linewidth}{@{\extracolsep{\fill}}lcccc@{}}
\toprule
& \multicolumn{2}{c}{\iclrrev{\textit{Study 1}}} & \multicolumn{2}{c}{\iclrrev{\textit{Study 2}}} \\
\cmidrule(lr){2-3}\cmidrule(lr){4-5}
\iclrrev{Metric} & \iclrrev{Open-ended} & \iclrrev{Binary} & \iclrrev{Interleaved} & \iclrrev{Batched} \\
\midrule
\iclrrev{Response time $\downarrow$} & \iclrrev{9.25 s} & \iclrrev{\textbf{2.69 s}} & \iclrrev{6.27 s} & \iclrrev{\textbf{5.28 s}} \\
\iclrrev{Preference $\uparrow$} & \iclrrev{0/20} & \iclrrev{\textbf{20/20}} & \iclrrev{6/20} & \iclrrev{\textbf{14/20}} \\
\bottomrule
\end{tabular*}
}

\vspace{-8pt}
\end{wraptable}

Human users favored the binary-question condition. In Study~1, all 20 participants preferred and responded faster in the binary condition, while 89.1\% of Korean open-ended responses contained at most three words. Compared with the lengthy MLLM-simulated open-ended responses in Table~\ref{tab:user_burden}, this reveals a simulator-human verbosity mismatch in open-ended interaction. Binary questions largely avoid this mismatch because simulated and real users share the same constrained response format. This makes binary interaction better aligned between simulated evaluation and real-user deployment. In Study~2, 14 of 20 participants (70\%) preferred batched delivery, consistent with prior work on interruption costs and notification batching~\citep{bailey2006attention,fitz2019batching}. See Appendix~\ref{app:user_study} for human evaluation details.

\newpage
\section{Analysis}

\subsection{\iclrrev{Termination-Step Analysis}}

\begin{wrapfigure}{r}{0.48\linewidth}
    \centering
    \vspace{-17pt}
    \setlength{\abovecaptionskip}{1pt}
    \setlength{\belowcaptionskip}{0pt}
    \includegraphics[width=\linewidth]{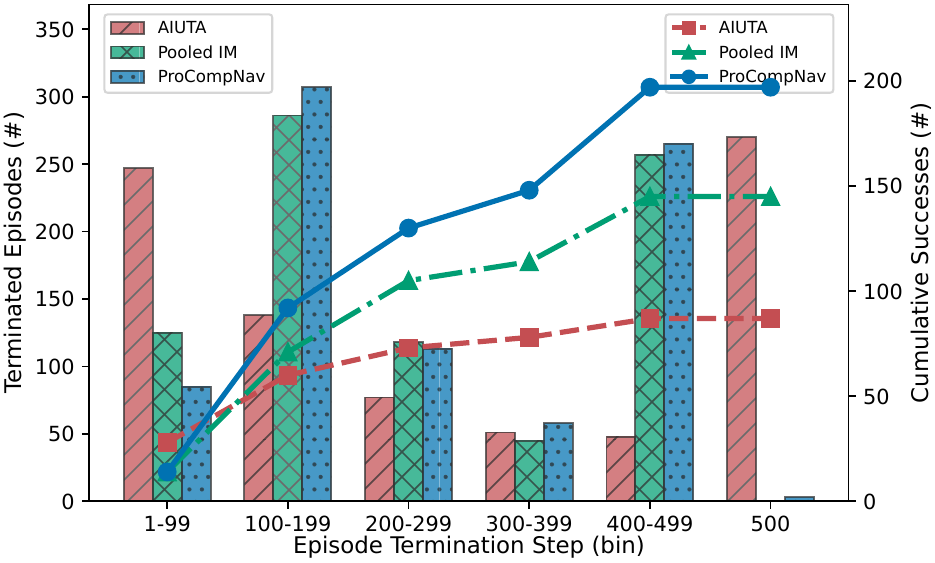}
    \vspace{-16pt}
    \caption{\iclrrev{Termination-step analysis of AIUTA, Pooled Independent Matching, and ProCompNav. Bars show terminated episodes in 100-step bins on the left axis, and lines show cumulative successful episodes on the right axis. The final bin denotes the 500-step horizon.}}
    \vspace{-10pt}
    \label{fig:stepbin_hist_cum_sr_total}
\end{wrapfigure}

To demonstrate the advantage of our collect-then-compare strategy, we compare the episode termination steps and success rates of AIUTA, \iclrrev{Pooled Independent Matching,} and \OursName on the CoIN-Bench Val Seen split in Figure~\ref{fig:stepbin_hist_cum_sr_total}.
AIUTA, which adopts an independent matching strategy, either makes a premature wrong target decision upon encountering a distractor (see the 1-99 step bin in Figure~\ref{fig:stepbin_hist_cum_sr_total}), or fails to make any target decision until the episode horizon is exhausted (see the 500 (max) step bin in Figure~\ref{fig:stepbin_hist_cum_sr_total}), both of which lead to a low success rate.
\iclrrev{Pooling reduces early commitments, while comparative judgment improves decisions after candidate collection. Relative to AIUTA, Pooled Independent Matching shifts terminations away from the first 100 steps. ProCompNav has a similar termination distribution to Pooled Independent Matching but accumulates more successes.}

\subsection{\iclrrev{Failure Analysis}}
\label{sec:failure_analysis}
\label{app:pruning_failure}

\begin{figure}[!h]
\vspace{-12pt}
\centering
\begin{minipage}[t]{0.53\textwidth}
\vspace{2\baselineskip}
\centering
\includegraphics[width=\linewidth]{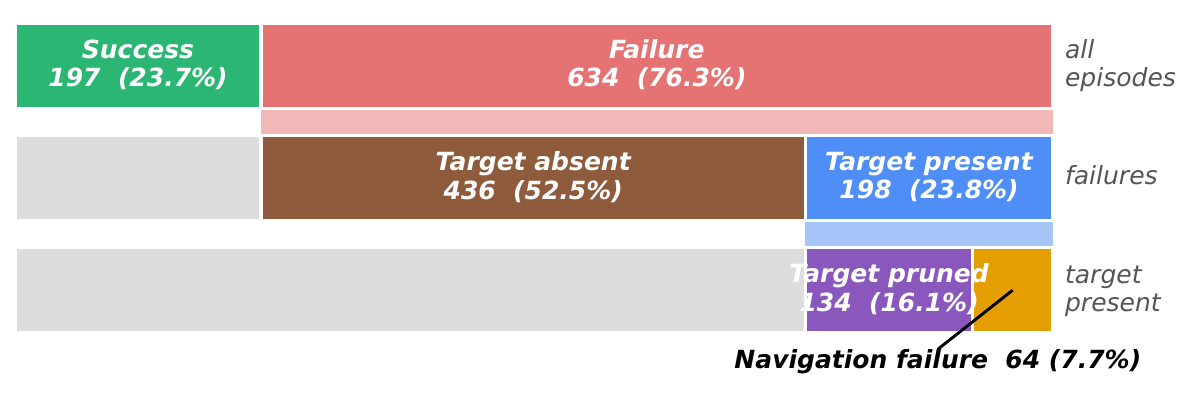}
\caption{\iclrrev{Failure breakdown on CoIN-Bench Val Seen. Percentages are over all 831 episodes.}}
\label{fig:tom3_failure_modes}
\end{minipage}\hfill
\begin{minipage}[t]{0.44\textwidth}
\vspace{0pt}
\captionof{table}{\iclrrev{Possible causes of target pruning on CoIN-Bench Val Seen ($n=134$). Shares are within this subset.}}
\label{tab:rebuttal_pruning_failure}
\vspace{2pt}
\begingroup
\centering\small
\setlength{\tabcolsep}{3pt}
\begin{tabular}{lrr}
\toprule
\iclrrev{Category} & \iclrrev{$n$} & \iclrrev{Share (\%)} \\
\midrule
\iclrrev{Caption Hallucination} & \iclrrev{24} & \iclrrev{17.9} \\
\iclrrev{Caption Omission} & \iclrrev{48} & \iclrrev{35.8} \\
\iclrrev{Core-set False Positive} & \iclrrev{24} & \iclrrev{17.9} \\
\iclrrev{Refinement False Negative} & \iclrrev{3} & \iclrrev{2.2} \\
\iclrrev{Refinement False Positive} & \iclrrev{0} & \iclrrev{0.0} \\
\iclrrev{Simulator-Judge Mismatch} & \iclrrev{35} & \iclrrev{26.1} \\
\midrule
\iclrrev{Total} & \iclrrev{134} & \iclrrev{100.0} \\
\bottomrule
\end{tabular}
\endgroup

\end{minipage}
\vspace{-10pt}
\end{figure}

\iclrrev{Figure~\ref{fig:tom3_failure_modes} shows that target absence is the dominant failure mode, accounting for 436 of 634 failures (68.8\%). These cases occur during VLFM-based exploration and object detection, before RCJ. Among the 198 target-present failures, the agent selects the target but fails during navigation in 64 episodes. In the remaining 134, the target is pruned during RCJ.}

\iclrrev{To investigate why targets are pruned during RCJ, we examine these 134 episodes using Gemini~3 Flash Preview\footnote{\label{fn:gemini_flash}\url{https://blog.google/products-and-platforms/products/gemini/gemini-3-flash/}} for post-hoc judgments (Table~\ref{tab:rebuttal_pruning_failure}). The dominant factors (53.7\%) are hallucination and attribute omission during caption generation from each candidate's multi-view images in the Pool Construction Stage. In 17.9\% of cases, the target is in the core set, but Gemini judges that it does not have the selected attribute. A \textbf{No} answer then removes the core set, including the target. Only 2.2\% are categorized as refinement false negatives, with no false positives observed at the conservative threshold $\tau=0.9$. A further 26.1\% involve Qwen3-VL-8B giving a different answer from Gemini's judgment. Even for a simple yes/no question, answers can depend on subjective interpretation (e.g., whether a shade is considered light brown). Appendix~\ref{app:failure_attribution} details the analysis procedure.}

\subsection{Qualitative Analysis}

\begin{figure}[H]
\centering
\includegraphics[width=0.85\linewidth, trim=0 0 0 0, clip]{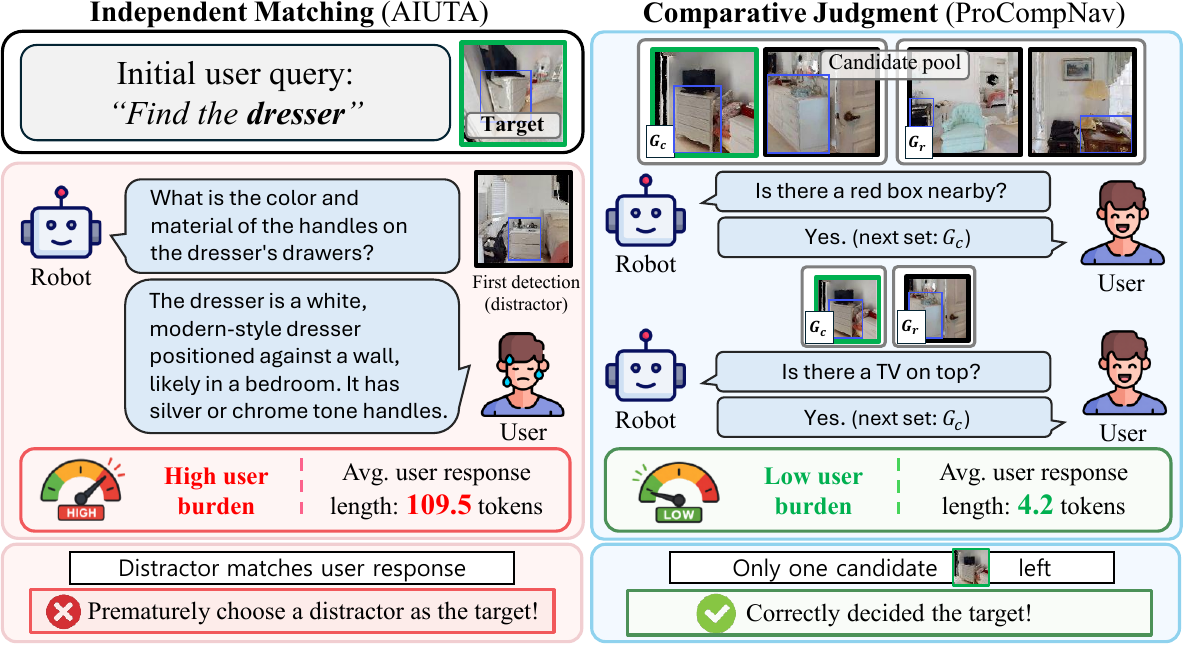}
\caption{Qualitative comparison of Independent Matching and Comparative Judgment under a minimally specified query (``Find the dresser''). \textbf{Independent Matching} (left) asks the user to describe an attribute of the intended target without knowing whether that attribute distinguishes it from distractors. This burdens the user with a long response and can lead the robot to mistakenly match a distractor sharing the same attribute. In contrast, \textbf{Comparative Judgment} (right) asks about discriminative attributes in a way that elicits only short user responses, reducing verbal burden while correctly isolating the user-intended target.}
\label{fig:qualitative_analysis}
\end{figure}

Figure~\ref{fig:qualitative_analysis} contrasts independent matching and \OursName on the ambiguous query ``Find the dresser''.
The independent matching baseline~\citep{taioli2025collaborative} (left) asks an open-ended attribute question (``What is the color and material of the handles\dots''), forcing a long descriptive response.
Even with this detailed answer, the baseline often decides on the first detected distractor as the target.
\OursName (right) instead defers its decision, builds a candidate pool, and asks discriminative binary questions.
In Round~1, ``Is there a red box nearby?'' splits the pool into a core set ($G_c$) and a remainder set ($G_r$), and a single Yes eliminates three distractors.
In Round~2, ``Is there a TV on top?'' isolates the true target ($T^*$).

\section{Conclusion}\label{sec:conclusion}

We studied proactive clarification in interactive instance navigation under ambiguous user requests.
We proposed \OursName, which first constructs a candidate pool and then identifies the target through Recursive Comparative Judgment.
Comparing candidates lets the robot select both an attribute and a value to split the pool, without requiring an attribute unique to the target.
This makes binary questions useful for eliminating candidates, beyond merely shortening responses.
On CoIN-Bench, \OursName improves Success Rate while reducing response length, and it achieves the highest Success Rate among evaluated methods on non-interactive TextNav.
In two human studies, participants preferred binary questions to open-ended questions and batched delivery to interleaved interaction.
Together, these results suggest that agentic robots should be designed for both task performance and user-friendly interaction.

\paragraph{Limitations}
ProCompNav is currently evaluated in simulation, and extending it to real-world robotic platforms remains an important future direction. \rebuttalrev{ProCompNav relies on its exploration and detection modules to place the target in the candidate pool, leaving room for improvement with stronger components. It currently does not recover the target once it is mistakenly pruned by incorrect binary feedback, motivating future work on recovery mechanisms.} Finally, requiring a minimum pool size may be inefficient when few same-category instances exist, motivating dynamic adaptation.

\newpage
\begin{ack}
This work was supported by the Institute of Information \& Communications Technology Planning \& Evaluation (IITP) grants funded by the Korea government (MSIT) (RS-2019-II191906, Artificial Intelligence Graduate School Program (POSTECH); IITP-2026-RS-2024-00437866, Information Technology Research Center (ITRC); IITP-2026-RS-2026-25616370, AI Star Fellowship Support Program), and by a Korea Institute for Advancement of Technology (KIAT) grant funded by the Ministry of Trade, Industry and Energy (MOTIE), Republic of Korea (RS-2025-00564342).
The authors declare no competing interests.
\end{ack}

\bibliographystyle{plainnat}
\bibliography{main}

\appendix
\clearpage
\section{Baselines}
For CoIN-Bench~\citep{taioli2025collaborative}, we compare \OursName against trained policies \textbf{Monolithic-GOAT}~\citep{khanna2024goat} and \textbf{PSL}~\citep{sun2024prioritized}.
Among training-free methods, \textbf{3D-Mem}~\citep{yang20253d} and \textbf{Context-Nav}~\citep{jang2026context} receive a detailed target description---a richer input than the target category available to \OursName---while \textbf{VLFM}~\citep{yokoyama2024vlfm} and \textbf{AIUTA}~\citep{taioli2025collaborative} receive only the target category.
VLFM navigates without interaction, whereas AIUTA asks the user simulator open-ended clarifying questions.
We further evaluate on TextNav to verify that \OursName generalizes to the non-interactive, detailed-description setting, using the same baselines as CoIN-Bench and additionally including \textbf{UniGoal}~\citep{yin2025unigoal}.
We include \textbf{AIUTA$^{*}$}, our reproduction of AIUTA with the same LLM and MLLM as \OursName; on TextNav, it is additionally adapted for the non-interactive setting (see Appendix~\ref{app:non-interactive}).
\noindent\textbf{Pooled independent matching} is a baseline we implement, pairing \OursName's pool construction stage with AIUTA's independent matching.
Once the pool is collected, the agent visits the candidates in two passes. 
In the first pass, it asks the user one open-ended question per candidate, conditioning each question on the facts accumulated from the previous candidates so that each new question elicits information not yet gathered.  
In the second pass, an LLM scores how consistent each candidate’s description is with the final fact set.
When multiple candidates are tied at the top score, an LLM is given their descriptions and asked to select the one most consistent with the accumulated facts.

\section{Pool Construction Stage Details}\label{app:pool_construction_details}

\noindent\textbf{Cross-view instance assignment.}
For each new detection of category-$c$, the agent back-projects the bounding box to a 3D point cloud and computes the symmetric overlap ratio against every existing candidate's accumulated point cloud $\mathcal{P}_i$, defined as the fraction of points in either cloud that have a neighbor in the other within radius $\epsilon = 0.03$\,m (we take the maximum of the two directed fractions).
If the maximum overlap is at least $0.3$, the detection is assigned to the matching candidate and its RGB view is added to $\mathcal{V}_i$. Otherwise, a new candidate is initialized.

\begin{figure}[!t]
\centering
\includegraphics[width=\linewidth]{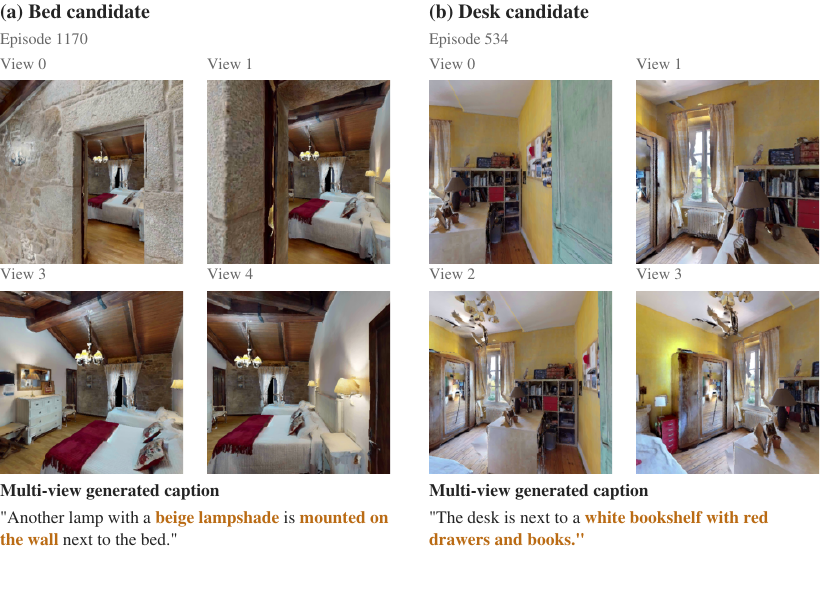}
\caption{\iclrrev{Multi-view observations and generated captions. Each panel shows recorded views of one candidate and part of its multi-view generated caption. Highlighted words describe details visible in the images. The generated captions are used as input to RCJ.}}
\label{fig:figure_final_merged_views}
\end{figure}

\noindent\textbf{Diverse view selection.}
For each candidate $i$, we extract a visual embedding for every view $v \in \mathcal{V}_i$ using DINOv2~\citep{oquab2023dinov2}, cluster the embeddings into $K = 6$ clusters via $K$-means, and select the view closest to each centroid as a representative.
We arrange the resulting representatives into the multi-view collage $I_i$.
\iclrrev{An MLLM filters out views in which the candidate is not clearly visible to reduce noise in caption generation.}
\iclrrev{To base RCJ's visual similarity on a clear observation, the MLLM selects the clearest retained view for the image embedding.}
\iclrrev{Figure~\ref{fig:figure_final_merged_views} connects recorded candidate views to generated descriptions used as RCJ input.}


\section{NLI-based Entailment Scoring}\label{app:entailment}
We instantiate the attribute verifier with a pretrained Natural Language Inference (NLI) classifier. For each instance description $d_i$ (premise) and candidate attribute $a$, we form the hypothesis ``the instance has attribute $a$'' and query the NLI model to obtain logits $\ell_E(d_i,a)$, $\ell_N(d_i,a)$, and $\ell_C(d_i,a)$ for entailment, neutral, and contradiction, respectively. We convert these logits into a scalar entailment score
\begin{align*}
s(d_i,a) = \sigma\!\big(\ell_E(d_i,a) - \max(\ell_N(d_i,a), \ell_C(d_i,a))\big) \;,
\end{align*}
where $\sigma$ is the sigmoid function. This score provides a calibrated measure of attribute support that is comparable across instances and attributes. In our implementation, we use a lightweight NLI backbone (\textit{e.g.}, DeBERTa-v3-large~\citep{he2021debertav3,laurer2024less}) and apply the same scoring to all candidates in $G_c$ and $G_r$.

\begin{figure}[!t]
\centering
\includegraphics[width=\linewidth, trim=0 0 0 0, clip]{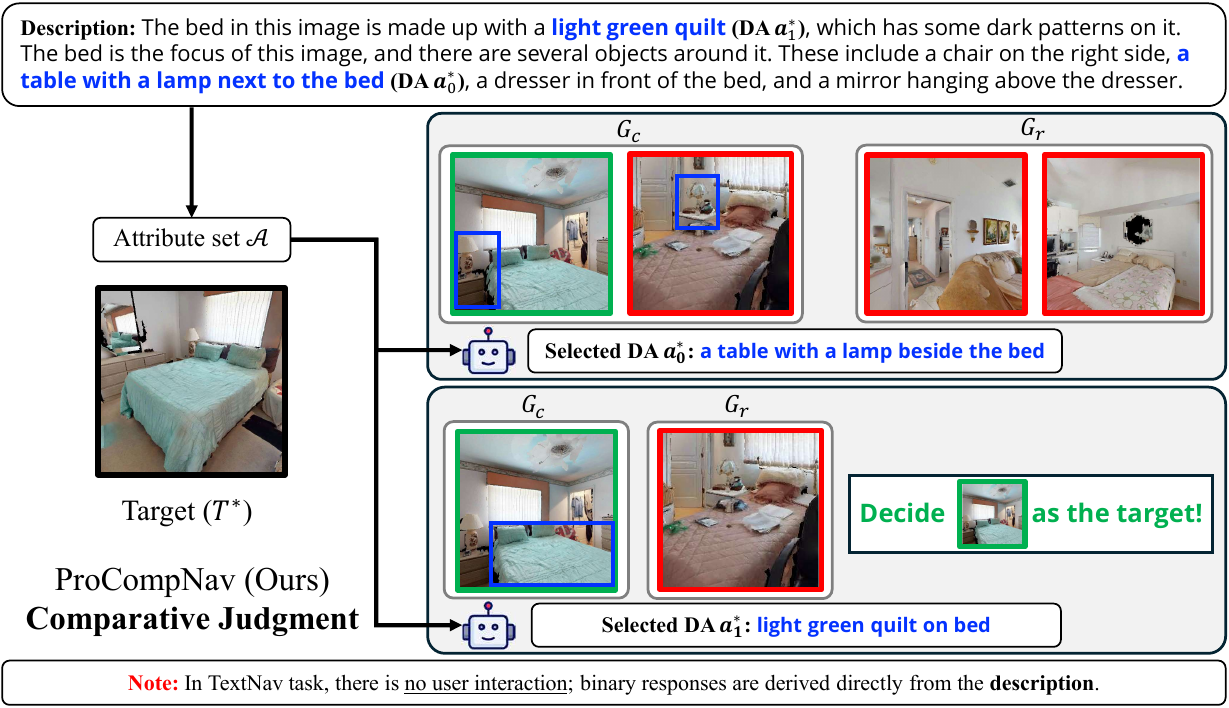}
\caption{TextNav adaptation of the Recursive Comparison Stage.
In TextNav, ProCompNav pre-extracts an attribute set $\mathcal{A}$ from the text goal before the Recursive Comparison Stage and, at each round, selects the discriminative attribute that best supports the current core set $G_c$. Since these attributes are derived from the goal, the active set is always updated to $G_c$. The agent terminates once the target-selection condition is met.}
\label{fig:qualitative_analysis_textnav}
\end{figure}

\section{Adaptation of ProCompNav to TextNav}\label{app:ours_textnav}
Unlike CoIN-Bench, TextNav provides a detailed target description at the beginning of each episode and does not allow user interaction. We use $N_{\min}=3$ on TextNav and start the Recursive Comparison Stage at step 600 if it has not started earlier. We disable the question-asking module and adapt the Recursive Comparison Stage in the following two ways.

\noindent\textbf{(1) Goal-derived attribute set.}
Instead of extracting attribute candidates from the current core set $G_c$ at each iteration, we parse a fixed attribute set $\mathcal{A}$ once from the text goal describing the target instance using the LLM. Since $\mathcal{A}$ is derived from the goal description, its attributes are assumed to hold for the target. Accordingly, after attribute-guided group refinement, we always set $U_{t+1}=G_c$, rather than choosing between $G_c$ and $G_r$ based on user feedback as in CoIN.

\noindent\textbf{(2) Candidate verification without user feedback.}
When $G_r=\emptyset$, we invoke a text-only LLM-based verifier. Specifically, if $|U_t|=1$, we verify the sole remaining candidate instance; if $G_r=\emptyset$ and $|G_c|\geq 2$, we verify the candidate instance $i \in G_c$ with the highest entailment score $s(d_i,a_t^*)$. The verifier compares the candidate description $d_i$ with the text goal describing the target instance and returns a binary accept/reject decision. If accepted, the agent executes \texttt{stop}; otherwise, it resumes the Pool Construction Stage to collect additional candidates from unexplored regions.


\section{Adapting Object Navigation for Instance-Level Exploration} \label{app:exploration}

Language-driven instance navigation methods often adopt an object navigation (ON) backbone for exploration---\textit{e.g.}, AIUTA~\citep{taioli2025collaborative} uses VLFM~\citep{yokoyama2024vlfm} and UniGoal~\citep{yin2025unigoal} uses SG-Nav~\citep{yin2024sg}. 
These methods repeatedly select the frontier with the highest reasoning score and are designed to reach \textit{any} instance of the target category. 
In instance navigation, however, the first observed instance is not necessarily the target, so the agent may need to continue exploring even after an initial detection.

Directly extending ON backbones to this setting introduces two practical issues. 
First, greedy top-1 frontier selection can repeatedly favor an unreachable or suboptimal frontier, causing local loops and reducing exploration coverage. 
Second, with a forward-facing monocular camera and a narrow field of view (30$^\circ$), the agent may pass nearby objects without observing them, leaving potential candidates undetected. 
To address these issues, we add two lightweight heuristics to our VLFM-based exploration backbone.

\subsection{Loop detection and frontier escape}


We maintain an exponential moving average (EMA) of the agent's position to detect trajectory stagnation. 
Let $\mathbf{p}_t$ denote the agent's position at step $t$. The trajectory center $\mathbf{c}_t$ and spread $s_t$ are updated as:
\begin{align*}
    \mathbf{c}_t &= \alpha\,\mathbf{p}_t + (1-\alpha)\,\mathbf{c}_{t-1}\;,\\
    s_t &= \alpha\,\|\mathbf{p}_t - \mathbf{c}_t\|^2 + (1-\alpha)\,s_{t-1}\;.
\end{align*}
We then define the loopness score as
\begin{equation*}
    \mathrm{loop}_t = \exp\!\left(-\frac{\|\mathbf{p}_t - \mathbf{c}_t\|^2}{s_t}\right)\;,
\end{equation*}
which approaches 1 when the agent remains near its recent trajectory center. 
The normalization by $s_t$ makes the score adaptive to the recent motion scale. 
If $\mathrm{loop}_t \geq 0.9$ persists for 5 consecutive steps while moving toward a frontier, we temporarily blacklist all frontiers in the same grid cell by assigning them a large negative score, forcing the planner to redirect exploration elsewhere.

\subsection{Line-of-sight rotation}

To compensate for the narrow field of view, we trigger an in-place 360$^\circ$ rotation when the surroundings are sufficiently open. 
We define the \textit{openness} of the current position $(x_0, y_0)$ as the fraction of 360 uniformly spaced angular bins (1$^\circ$ resolution) whose line of sight is not blocked on the occupancy map:
\begin{equation*}
    \mathrm{openness} = 1 - \frac{|\{\text{occluded bins}\}|}{360}\;.
\end{equation*}
A bin at angle $\theta$ is marked occluded if the ray in that direction intersects an obstacle, where each obstacle point $(x_i, y_i)$ is associated with angle $\theta_i = \mathrm{atan2}(y_i - y_0,\, x_i - x_0)$. 
We execute a 360$^\circ$ rotation when $\mathrm{openness} \geq 0.1$ and the agent is at least 1.0\,m away from the previous rotation point, which avoids redundant rotations in confined areas. 

\section{Adaptation of AIUTA to TextNav}\label{app:non-interactive}
For a fair comparison under the same model condition, we adapt AIUTA~\citep{taioli2025collaborative} to both CoIN-Bench~\citep{taioli2025collaborative} and TextNav~\citep{sun2024prioritized} using the same Qwen3-VL-8B~\citep{bai2025qwen3} MLLM as in our method. 
On CoIN-Bench, AIUTA* follows the original interactive setting. 
On TextNav, however, we remove the user-interaction module and replace the original dynamically updated fact memory with the given detailed text goal, which is kept fixed throughout the episode. 
Accordingly, when a target-category object is detected, AIUTA* no longer considers the \texttt{Ask} action; instead, it compares the detected object against the fixed text goal and chooses only between \texttt{Stop} and \texttt{Skip}. 
If the detected object is judged to match the detailed goal, the agent executes \texttt{stop}; otherwise, it skips the instance and continues exploration.

\noindent\begin{minipage}{\linewidth}
\section{\iclrrev{Question Format and Candidate Comparison}}
\label{app:binary_ablation}

\begingroup
\centering
\captionof{table}{\iclrrev{Question format and candidate comparison on all 831 CoIN-Bench Val Seen episodes, using the same pool construction stage and Qwen3-VL-8B backbone. Pooled Independent Matching (PIM) and Binary PIM accumulate target facts and score candidates independently. The ablation variants ``No candidate information'' and ``Single candidate'' prune candidates using binary feedback, as in ProCompNav's RCJ. MaxQ is the percentage of all episodes reaching the four-question limit. NQ averages over interactive episodes.}}
\label{tab:binary_ablation}
\begingroup
\small
\setlength{\tabcolsep}{2.5pt}
\renewcommand{\arraystretch}{1.15}
\iclrrev{\begin{tabularx}{\linewidth}{@{}>{\raggedright\arraybackslash}p{0.245\linewidth}>{\raggedright\arraybackslash}p{0.11\linewidth}>{\raggedright\arraybackslash}Xrrrrr@{}}
\toprule
Method & Question & Attribute selection & SR$\uparrow$ & SPL$\uparrow$ & RL$\downarrow$ & NQ$\downarrow$ & MaxQ$\downarrow$ \\
\midrule
\multicolumn{8}{l}{\textit{\textbf{Question Format (per-candidate questioning)}}} \\
Pooled Independent Matching (PIM) & Open-ended & Per candidate & 17.5 & 5.1 & 460.2 & 4.1 & - \\
Binary PIM & Binary & Per candidate & 17.2 & 5.2 & 7.5 & 4.3 & - \\
\midrule
\multicolumn{8}{l}{\textit{\textbf{Attribute Selection (binary questions)}}} \\
No candidate information & Binary & No candidate captions & 14.1 & 3.6 & 5.7 & 3.0 & 56 \\
Single candidate & Binary & One candidate caption & 19.6 & 5.7 & 5.2 & 2.7 & 31 \\
\textbf{ProCompNav} & Binary & Structured comparison & \textbf{23.7} & \textbf{7.0} & \textbf{4.2} & \textbf{2.2} & \textbf{2} \\
\bottomrule
\end{tabularx}}
\endgroup
\endgroup

\end{minipage}
\par\medskip

\noindent\textbf{\iclrrev{Question format.}} \iclrrev{We compare Pooled Independent Matching (PIM) and Binary PIM to verify the response-length benefit of binary questions. Both ask one question per candidate without a four-question limit and retain fact accumulation and independent candidate scoring. Binary PIM replaces PIM's open-ended questions with binary attribute-value questions. Its LLM receives only the current candidate's description and accumulated target facts, without other candidate descriptions. Simulated RL decreases from 460.2 to 7.5 tokens while SR remains comparable, confirming the response-length reduction from binary questions under the same per-candidate questioning strategy.}

\noindent\textbf{\iclrrev{Attribute selection.}} \iclrrev{We next vary how attribute-value pairs are selected for binary questions under the main setting's four-question budget (Section~\ref{sec:exp_setup}). ``No candidate information'' follows a twenty-questions-style interaction: the first question uses only the target category (e.g., sofa), and subsequent questions use previous questions and user responses. ``Single candidate'' additionally receives one candidate caption per question, selected sequentially as in PIM and Binary PIM. This is intended to avoid implicit candidate comparison that could arise from presenting all captions together. It takes the next unused caption among the remaining candidates in collection order. Both include previously queried attributes in the prompt for each new question and instruct the LLM not to reuse them.}

\iclrrev{After each answer, candidates inconsistent with it are pruned. The two variants use the same NLI verifier and entailment-score threshold ($0.9$) as ProCompNav, including filtering after re-exploration and entailment-score-based selection at the question limit. Questioning ends when one candidate remains or the four-question budget is exhausted. In the latter case, a candidate is selected using entailment scores. If no candidates remain after pruning, we apply the re-exploration and fallback rules. Thus, unlike per-candidate questioning, NQ depends on how effectively each question reduces the pool. These ablations examine how attribute-value choices that do not effectively narrow the candidate pool affect RL and NQ, even with binary responses.}

\noindent\textbf{\iclrrev{Beyond Binary Format.}} \iclrrev{Binary questions can greatly reduce response length relative to open-ended questions, but this format alone does not ensure fewer questions. ``Single candidate'', which receives one candidate caption per question, improves SR and reduces NQ to 2.7 and MaxQ to 31\% compared with ``No candidate information''. However, without identifying a value that clearly eliminates candidates, binary questioning can still require many questions to cover possible values. ProCompNav compares candidates to choose both the attribute and the value to verify, achieving higher SR with NQ of 2.2 and MaxQ of 2\% (Table~\ref{tab:binary_ablation}).}

\section{\iclrrev{Increasing Query Specificity}}\label{app:query_specificity}

\begin{table}[H]
\centering\color{black}\small
\caption{Performance with increasingly specific CoIN queries. $k$ is the number of target hints. \iclrrev{All results use the 621 Val Seen episodes whose target descriptions yield at least three parsed atomic hints. $\dagger$The $k=0$ reference is also aggregated over these 621 episodes, rather than the full Val Seen split.}}
\label{tab:rebuttal_hints}
\resizebox{\linewidth}{!}{%
\begin{tabular}{lcccccccccccc}
\toprule
& \multicolumn{3}{c}{$k=0$\textsuperscript{$\dagger$}} & \multicolumn{3}{c}{$k=1$} & \multicolumn{3}{c}{$k=2$} & \multicolumn{3}{c}{$k=3$} \\
\cmidrule(lr){2-4}\cmidrule(lr){5-7}\cmidrule(lr){8-10}\cmidrule(lr){11-13}
Method & SR $\uparrow$ & RL $\downarrow$ & NQ $\downarrow$ & SR $\uparrow$ & RL $\downarrow$ & NQ $\downarrow$ & SR $\uparrow$ & RL $\downarrow$ & NQ $\downarrow$ & SR $\uparrow$ & RL $\downarrow$ & NQ $\downarrow$ \\
\midrule
AIUTA$^*$ & \iclrrev{12.2} & \iclrrev{116.1} & \iclrrev{1.3} & 10.1 & 16.2 & 2.0 & 6.8 & 5.5 & 1.1 & 8.5 & 8.9 & 1.2 \\
Pooled & \iclrrev{16.6} & \iclrrev{430.7} & \iclrrev{3.9} & 19.3 & 421.1 & 3.8 & 17.2 & 455.9 & 4.0 & 19.2 & 452.6 & 4.1 \\
\textbf{ProCompNav} & \iclrrev{\textbf{21.9}} & \iclrrev{\textbf{4.1}} & 2.2 & \textbf{21.9} & \textbf{4.2} & 2.3 & \textbf{21.7} & \textbf{4.3} & 2.3 & \textbf{22.4} & \textbf{4.2} & 2.3 \\
\bottomrule
\end{tabular}}
\end{table}

\iclrrev{We use Qwen3.6-Plus~\citep{qwen36plus} for atomic hint extraction.}
\iclrrev{We evaluate all $k\in\{0,1,2,3\}$ on the same 621 Val Seen episodes whose target descriptions yield at least three parsed atomic hints. We augment the category-only CoIN query with $k$ hints, which ProCompNav applies before extracting new DAs and querying the user. ProCompNav achieves the highest SR at every $k$, but SR and NQ do not improve monotonically as hints are added. An atomic target fact may also hold for distractors, so more hints do not necessarily resolve ambiguity.}

\begin{figure}[!htbp]
\centering
\includegraphics[width=\linewidth]{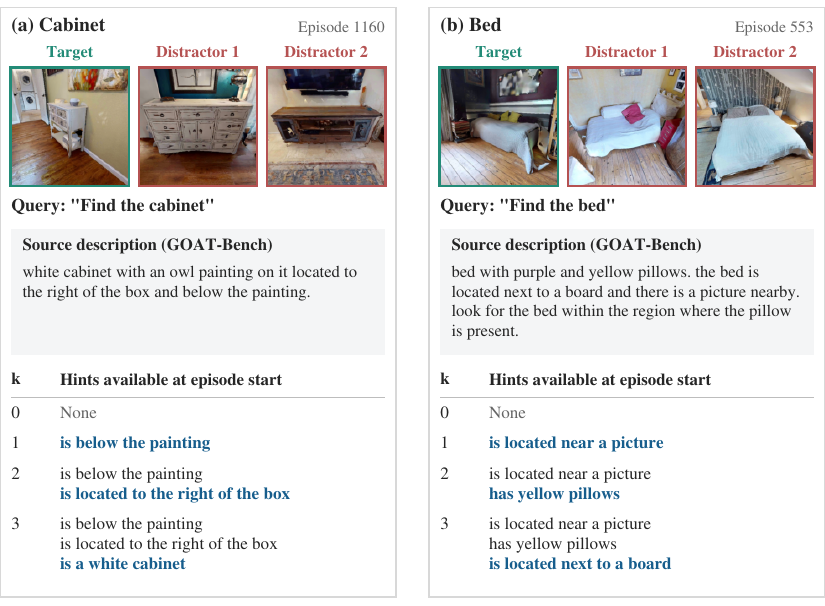}
\caption{\iclrrev{Atomic hints extracted from GOAT-Bench~\citep{khanna2024goat} target descriptions provided in CoIN-Bench. Gray boxes reproduce the source descriptions. Each hint specifies one target property or relation. The category query is unchanged for all $k$; hints are supplied separately. Each row retains the previous hints and adds the bold blue hint. Hint wording is preserved. Green and red borders indicate the target and two selected same-category distractors per episode. Images are re-rendered for illustration, not agent observations.}}
\label{fig:atomic_hint_examples}
\end{figure}

\iclrrev{Figure~\ref{fig:atomic_hint_examples} illustrates the nested hint inputs alongside their source target descriptions. The category query stays fixed, and the separate hint list grows by one atomic statement at each $k$.}

\par\medskip
\noindent\begin{minipage}{\linewidth}
\section{\iclrrev{Core-Set Selection}}
\label{app:core_set_ablation}
\begingroup
\centering
\captionof{table}{\iclrrev{Core-set selection on CoIN-Bench Val Seen. NQ averages questions over episodes with interaction, and MaxQ denotes the percentage of all episodes reaching the four-question limit. Stage~2 time is the average total time spent in the Recursive Comparison Stage per episode, excluding pool construction and navigation.}}
\label{tab:core_set_ablation}
\begingroup
\small
\setlength{\tabcolsep}{4pt}
\renewcommand{\arraystretch}{1.15}
\iclrrev{\begin{tabularx}{\linewidth}{@{}Xrrrrrr@{}}
\toprule
Method & SR$\uparrow$ & SPL$\uparrow$ & RL$\downarrow$ & NQ$\downarrow$ & MaxQ$\downarrow$ & Stage~2 time$\downarrow$ \\
\midrule
w/o core set (KMeans) & 20.6 & 5.8 & 4.5 & 2.4 & 9\% & 10.8\,s \\
\textbf{ProCompNav} & \textbf{23.7} & \textbf{7.0} & \textbf{4.2} & \textbf{2.2} & \textbf{2\%} & \textbf{6.3\,s} \\
\bottomrule
\end{tabularx}}
\endgroup
\endgroup

\end{minipage}
\par\medskip

\noindent\textbf{\iclrrev{Setup.}} \iclrrev{We replace core selection with KMeans ($k=2$, \texttt{random\_state=42}, \texttt{n\_init=10}), using the same 831 Val Seen episodes and RTX 3090 GPUs. KMeans concatenates separately $\ell_2$-normalized text and image embeddings, each scaled by $1/\sqrt{2}$, matching ProCompNav's equally weighted similarity. Pool construction, NLI attribute selection, refinement, the user simulator, and the four-question budget remain unchanged.}

\noindent\textbf{\iclrrev{Results.}} \iclrrev{Compared with KMeans using the same text and image features, ProCompNav achieves higher SR and lower RL, NQ, and MaxQ. It also takes less time in Stage~2 (Table~\ref{tab:core_set_ablation}).}

\section{Effect of The Refinement Threshold}\label{app:refinement_threshold}

\begin{table}[t]
\centering
\caption{Ablation on refinement threshold on Val Seen.}
\small
\setlength{\tabcolsep}{6pt}
\begin{tabular}{lcccc}
\toprule
Threshold & SR $\uparrow$ & SPL $\uparrow$ & RL $\downarrow$ & NQ $\downarrow$ \\
\midrule
No refinement & 21.8 & 6.0 & 4.2 & 2.1 \\
$\tau = 0.5$ & 22.0 & 6.3 & 4.1 & 2.2 \\
$\tau = 0.7$ & 22.6 & 6.5 & 4.2 & 2.2 \\
$\tau = 0.9$ (ours) & 23.7 & 7.0 & 4.2 & 2.2 \\
\bottomrule
\end{tabular}
\label{tab:refinement_threshold_ablation}
\end{table}

Table~\ref{tab:refinement_threshold_ablation} shows that refinement improves SR and SPL over the no-refinement variant with almost no increase in user burden.
Among the refinement thresholds, the conservative setting $\tau=0.9$ performs best, achieving the highest SR and SPL while keeping RL and NQ nearly unchanged.
This suggests that refinement is beneficial, but should be applied conservatively to avoid introducing unreliable comparisons.

\noindent\textbf{Core-set sizes.} \iclrrev{Across 1,752 grouping rounds, all 1,130 rounds with at least three candidates end in a core pair, while the remaining 622 two-candidate rounds use a singleton core. Refinement expands the core in 139 rounds, increasing its mean size from 2.00 to 2.13. The final partition contains two non-empty groups in 1,128 of the 1,130 peeling rounds.}

\begin{figure}[!htbp]
\centering
\includegraphics[width=\linewidth]{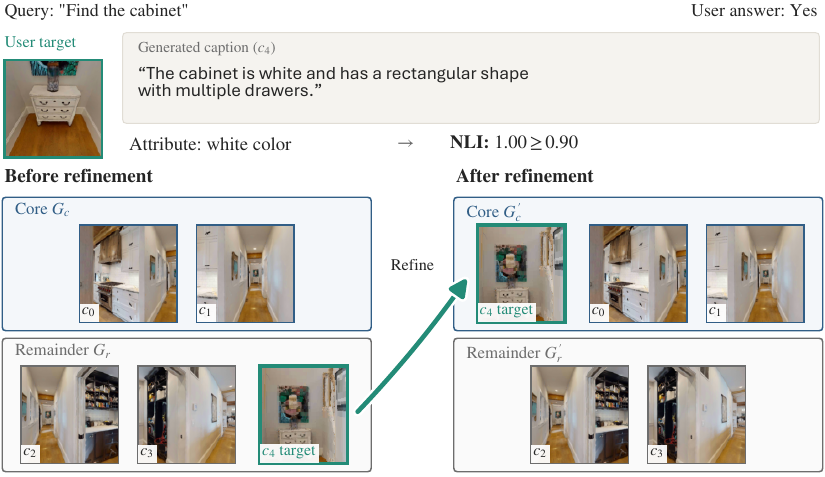}
\caption{\iclrrev{Attribute-guided group refinement (episode 1617). The quotation is the generated caption excerpt for $c_4$. Its support for ``white color'' exceeds the NLI threshold ($1.00\geq\tau=0.9$), moving the target from $G_r$ into $G_c'$. For the same recorded Yes answer, the unrefined partition would prune $c_4$, while the refined partition retains it. Candidate images are crops of recorded views, with a zoom for $c_4$; the user target reference is unavailable to the NLI verifier.}}
\label{fig:refinement_example}
\end{figure}

\noindent\textbf{\iclrrev{Refinement example.}} \iclrrev{Figure~\ref{fig:refinement_example} shows how refinement preserves a target initially placed in the remainder set despite sharing the selected attribute with the core.}

\section{Effect of the Pool Size Threshold}
\begin{figure}[t]
\centering

\begin{minipage}[c]{0.45\linewidth}
  \centering
  \includegraphics[width=\linewidth]{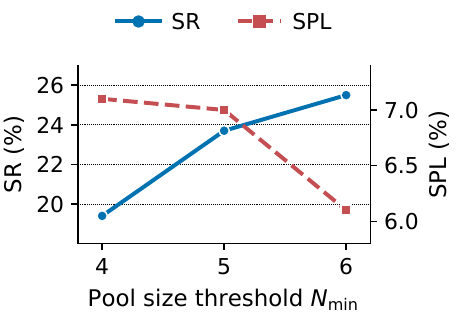}
\end{minipage}
\hfill
\begin{minipage}[c]{0.51\linewidth}
  \centering
  \begingroup
  \small
  \setlength{\tabcolsep}{3pt}
  \renewcommand{\arraystretch}{1.18}
  \begin{tabular*}{\linewidth}{@{\extracolsep{\fill}}crrrrr@{}}
    \toprule
    $N_{\min}$ & SR$\uparrow$ & SPL$\uparrow$ & RL$\downarrow$ & NQ$\downarrow$ & Steps$\downarrow$ \\
    \midrule
    4 & 19.4 & 7.1 & 3.8 & 2.0 & 212.7 \\
    \textbf{5}$^{\dagger}$ & 23.7 & 7.0 & 4.2 & 2.2 & 257.0 \\
    6 & 25.5 & 6.1 & 4.5 & 2.4 & 299.3 \\
    \bottomrule
  \end{tabular*}
  \par\smallskip
  {\footnotesize $\dagger$Default setting.}
  \endgroup
\end{minipage}

\caption{
  Effect of the candidate pool size threshold $N_{\min}$ that triggers the Recursive Comparison Stage on CoIN-Bench. (Left) SR and SPL under different $N_{\min}$ values. A larger $N_{\min}$ improves comparative judgment and thus increases SR, but leads to lower SPL. (Right) Detailed results for all evaluation metrics, including user burden and navigation cost. Based on this trade-off, we choose $N_{\min}=5$ as the default setting on CoIN-Bench.
}
\label{fig:ablation_n}
\end{figure}

In Fig.~\ref{fig:ablation_n}, we further analyze the effect of the candidate pool size threshold $N_{\min}$ that triggers the Recursive Comparison Stage on CoIN-Bench. When $N_{\min}=4$, the Recursive Comparison Stage is invoked earlier, yielding the highest SPL and fewest steps, but the candidate pool may not yet be sufficiently representative, resulting in a low SR of 19.4. Increasing $N_{\min}$ to 5 substantially improves SR to 23.7 (+4.3) while incurring only a modest SPL drop, suggesting that the additional exploration translates directly into SR gains by providing a more reliable comparison set. In contrast, $N_{\min}=6$ provides only a limited SR gain (+1.8) but noticeably worsens efficiency and interaction cost. We therefore choose $N_{\min}=5$ as the default setting on CoIN-Bench, as it offers the best balance between success rate, navigation efficiency, and user burden.

\subsection{\rebuttalrev{Navigation efficiency across target distance}}
\begin{table}[t]
\centering\color{black}\small
\caption{Path efficiency across target distances on Val Seen. \iclrrev{Distance is the geodesic distance from the agent's initial position to the target. No episode has a distance of $3$\,m or less. The three distance bins contain 25, 251, and 555 episodes, respectively. Results are obtained from separate evaluation runs using the same evaluation protocol as the main experiments.} Lower mean steps are better.}
\label{tab:rebuttal_path_efficiency}
\resizebox{\linewidth}{!}{%
\begin{tabular}{lccccc}
\toprule
Method & 3-5 m SR & 5-7 m SR & $>$7 m SR & Overall SR & Mean steps \\
\midrule
AIUTA$^*$ & 8.0 & 12.8 & 10.8 & \iclrrev{11.3} & 261.8 \\
Pooled & 28.0 & 22.3 & 13.3 & \iclrrev{16.5} & \textbf{242.3} \\
\textbf{ProCompNav} & \textbf{28.0} & \textbf{31.1} & \textbf{20.2} & \iclrrev{\textbf{23.7}} & 257.0 \\
\bottomrule
\end{tabular}}
\end{table}

\iclrrev{Table~\ref{tab:rebuttal_path_efficiency} shows that ProCompNav outperforms AIUTA$^*$ across all geodesic-distance groups with comparable mean steps. Pooled Independent Matching accounts for much of this gain, while comparative judgment adds larger improvements at longer distances.}

\begin{figure}[p]
  \centering
  \includegraphics[width=\linewidth]{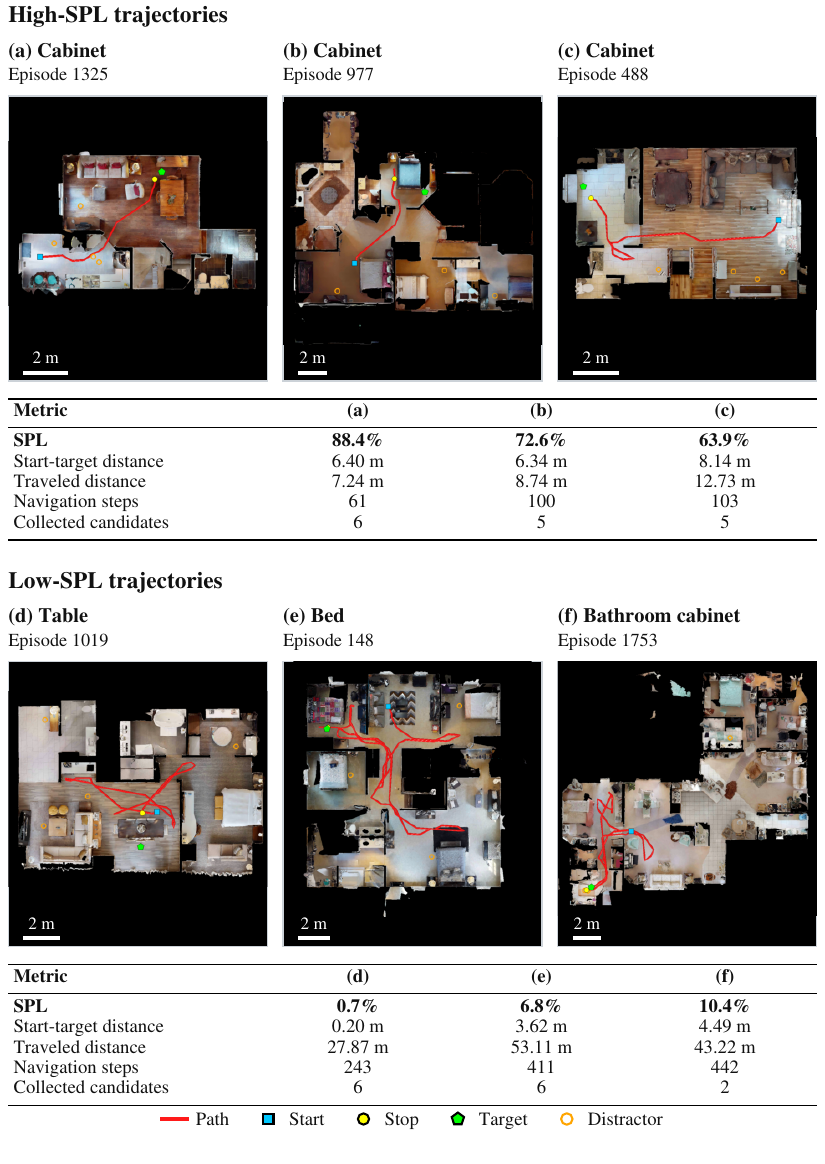}
  \caption{Successful navigation trajectories from a visualization rerun of 12 preselected CoIN-Bench Val Seen episodes. The top and bottom rows show the three highest- and lowest-SPL successes, with metrics aligned below each row. Red lines show recorded routes. Start-target distance is the geodesic distance from the agent's initial position to the nearest target goal viewpoint used in SPL evaluation. Traveled distance sums recorded movement during candidate collection and final navigation. Collected candidates are counted after merging at the first RCJ call. Maps are independently scaled, with metric scale bars.}
  \label{fig:trajectory-efficiency}
\end{figure}

\iclrrev{Figure~\ref{fig:trajectory-efficiency} shows the three highest- and three lowest-SPL successes in a visualization rerun. Delayed candidate collection can prolong exploration and produce detours. In (f), only two candidates have been collected by step 400, below $N_{\min}=5$. RCJ is then triggered by the step threshold, and the agent subsequently reaches the target.}

\section{\rebuttalrev{Stability Across Random Seeds}}\label{app:multiseed}
\iclrrev{For each method, we use master seeds 0, 1, and 2 to initialize Python's \texttt{random}, NumPy, PyTorch, and CUDA random-number states. We also set \texttt{config.habitat.seed} to the master seed and enable episode shuffling, so the episode order varies across seeds.}
\begin{table}[H]
\centering\color{black}\small
\caption{Evaluation with three seeds on Val Seen Synonyms. Values are mean $\pm$ standard deviation.}
\label{tab:rebuttal_multiseed}
\begin{tabular}{lcc}
\toprule
Method & SR (\%) & SPL (\%) \\
\midrule
AIUTA$^*$ & 15.1 $\pm$ 0.7 & 8.3 $\pm$ 0.3 \\
Pooled & 20.3 $\pm$ 2.7 & 6.5 $\pm$ 1.0 \\
\textbf{ProCompNav} & \textbf{27.5 $\pm$ 1.5} & \textbf{8.5 $\pm$ 0.4} \\
\bottomrule
\end{tabular}
\end{table}

\iclrrev{Across three seeds, ProCompNav achieves the highest mean SR and SPL. Its SR margin over Pooled Independent Matching is larger than either method's standard deviation (Table~\ref{tab:rebuttal_multiseed}).}

\section{\rebuttalrev{Results with Different MLLMs}}
\begin{table}[H]
\centering\color{black}\small
\caption{Results with different MLLMs on CoIN-Bench Val Seen Synonyms. SR is Success Rate.}
\label{tab:rebuttal_backbone}
\begin{tabular}{lcc}
\toprule
Method & Qwen3-VL-8B & Gemma-3-12B \\
\midrule
3D-Mem$^*$ & 18.9 & 12.0 \\
\textbf{ProCompNav} & \textbf{28.1} & \textbf{20.9} \\
\bottomrule
\end{tabular}
\end{table}

\iclrrev{ProCompNav outperforms 3D-Mem$^*$ with both Qwen3-VL-8B and Gemma-3-12B~\citep{team2025gemma}, showing that its advantage is not limited to one MLLM (Table~\ref{tab:rebuttal_backbone}). We evaluate both methods on the 359 Val Seen Synonyms episodes with the same MLLM in each comparison. Because 3D-Mem reports no CoIN-Bench result, 3D-Mem$^*$ is our reproduction.}

\clearpage

\section{Robustness Across User Simulators}
\begin{table}[!t]
\centering
\caption{
Effect of the user simulator on ProCompNav performance on CoIN-Bench.
We compare the original Qwen3-VL-8B user simulator with Gemini 3 Flash Preview.
}
\label{tab:user_simulator_comparison}
\resizebox{\linewidth}{!}{
\begin{tabular}{lcccccccccccc}
\toprule
\multirow{2}{*}{User Simulator}
& \multicolumn{4}{c}{Val Seen}
& \multicolumn{4}{c}{Val Seen Synonyms}
& \multicolumn{4}{c}{Val Unseen} \\
\cmidrule(lr){2-5}
\cmidrule(lr){6-9}
\cmidrule(lr){10-13}
& SR $\uparrow$ & SPL $\uparrow$ & RL $\downarrow$ & NQ $\downarrow$
& SR $\uparrow$ & SPL $\uparrow$ & RL $\downarrow$ & NQ $\downarrow$
& SR $\uparrow$ & SPL $\uparrow$ & RL $\downarrow$ & NQ $\downarrow$ \\
\midrule
Qwen3-VL-8B
& 23.7 & 7.0 & 4.2 & 2.2
& 28.1 & 8.5 & 4.3 & 2.2
& 17.0 & 6.2 & 4.2 & 2.3 \\
Gemini 3 Flash Preview
& 23.1 & 6.7 & 4.1 & 2.1
& 29.0 & 8.5 & 4.1 & 2.1
& 15.5 & 5.3 & 4.0 & 2.1 \\
\bottomrule
\end{tabular}
}
\end{table}

\iclrrev{We also evaluate ProCompNav using Gemini 3 Flash Preview as an alternative user simulator (Table~\ref{tab:user_simulator_comparison}). The results across SR, SPL, RL, and NQ are similar overall to those with Qwen3-VL-8B.}


\section{Computational Cost}\label{app:computational_cost}
\begin{table}[t]
\centering
\caption{Per-episode computational cost comparison between AIUTA* and ProCompNav on the Val Seen Synonyms split (359 episodes). Although ProCompNav takes more navigation steps to proactively collect candidates, it reduces MLLM/LLM inference time by 63.8\% relative to AIUTA*, replacing repeated open-ended MLLM scoring with lightweight NLI verification, and thus achieves a lower total wall-clock time.}
\label{tab:compute_cost_comparison}
\small
\setlength{\tabcolsep}{6pt}
\begin{tabular}{lcc}
\toprule
Metric & AIUTA* & ProCompNav \\
\midrule
Total runtime (s)               & 707.74  & 549.60 \\
Number of steps                 & 203.3  & 271.5 \\
\midrule
MLLM/LLM inference time (s)     & 229.01  & 82.89 \\
NLI inference time (s)          & -      & 1.29 \\
Exploration time (s)            & 478.72  & 465.42 \\
\bottomrule
\end{tabular}
\end{table}

Table~\ref{tab:compute_cost_comparison} breaks down the per-episode computational cost. All computational costs were measured on two NVIDIA RTX 3090 GPUs with 24GB memory each. The dominant saving comes from MLLM/LLM inference: AIUTA* invokes the MLLM repeatedly to score each candidate independently and to generate open-ended questions, accumulating 229.01s per episode, whereas ProCompNav concentrates its MLLM calls in the Pool Construction Stage's captioning and a small number of attribute extractions in the Recursive Comparison Stage, totaling only 82.89s. The NLI verifier used for DA scoring adds a negligible 1.29s.

\section{Instance Merge Analysis}\label{app:instance_merge}

\begin{figure}[!t]
\centering
\includegraphics[width=\linewidth]{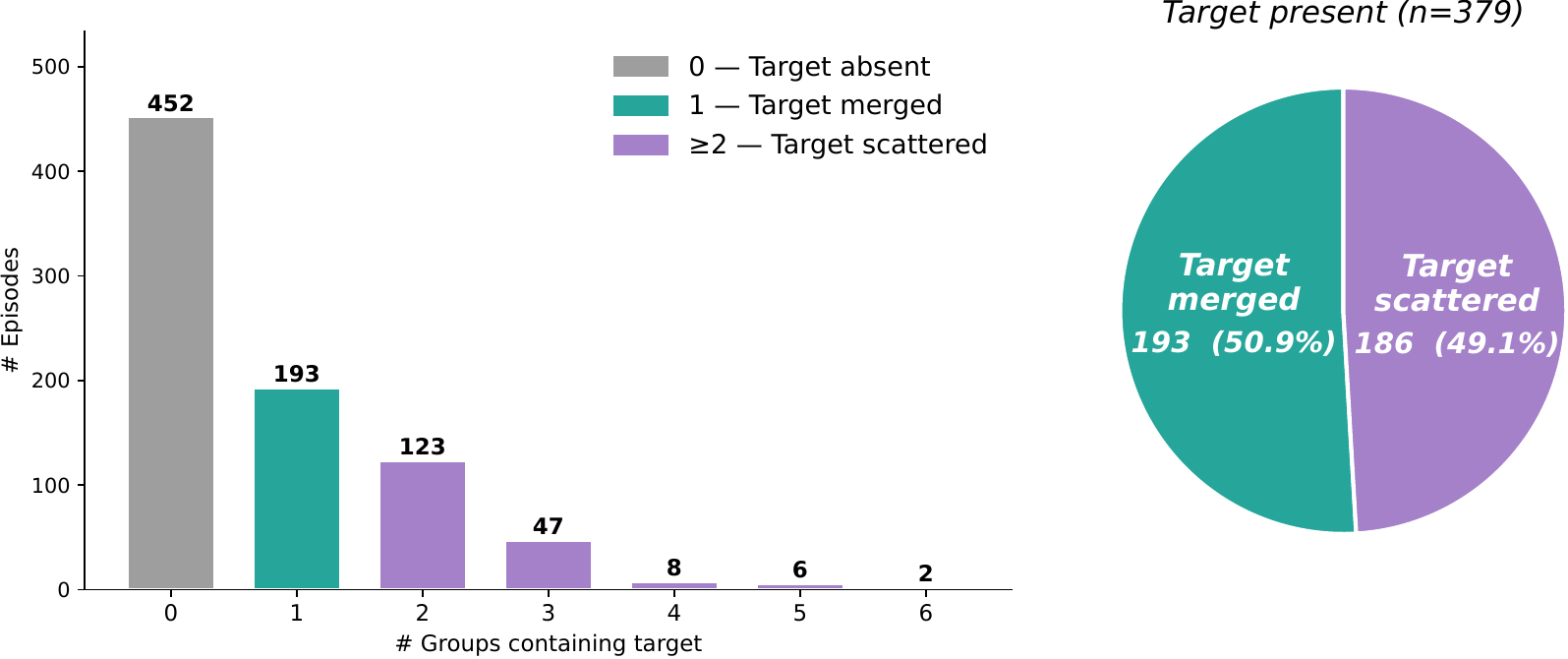}
\caption{
Instance merge quality of the Pool Construction stage on the CoIN-Bench Val Seen split\iclrrev{, assessed using VLM judgments of candidate collages}.
\textbf{Left}: distribution of the number of candidates whose multi-view collage contains the target. \textit{Target absent} (gray) marks episodes where no candidate contains the target. \textit{Target merged} (teal) marks episodes where the target's views are fully consolidated into a single candidate. \textit{Target scattered} (purple) marks episodes where the target's views are split across two or more candidates.
\textbf{Right}: among the 379 episodes whose pool \iclrrev{is judged to contain} the target, the shares with the target merged and scattered.
}
\label{fig:tom3_instance_merge}
\end{figure}

The Pool Construction Stage attempts to consolidate observations of the same physical instance from multiple views into a single candidate $c_i$; we assess how reliably this consolidation succeeds for the target instance.
For each episode in CoIN-Bench Val Seen, we feed the multi-view collage $I_i$ of every candidate to a high-performance VLM oracle (Gemini~3 Flash Preview\footref{fn:gemini_flash}) and label each candidate as containing the target or not.
\iclrrev{These VLM judgments can miss the target even in successful episodes, so target-presence counts can differ from the failure analysis.}
Among the 379 episodes whose pool \iclrrev{is judged to contain} the target (Fig.~\ref{fig:tom3_instance_merge}), 50.9\% have the target's views consolidated into a single candidate (\textit{Target merged}).
In the remaining 49.1\% the target's views are split across two or more candidates (\textit{Target scattered}).

This incomplete consolidation harms downstream behavior in two ways.
First, the multi-view description $d_i$ is generated from a partial set of views per candidate, so the description used by the Recursive Comparison Stage misses information that other views of the same instance carry.
Second, when the same target instance is treated as multiple distinct candidates, the Recursive Comparison Stage poses unnecessary discriminative questions to the user.
Improving cross-view consolidation in the Pool Construction Stage is therefore a direct lever for both caption quality and user efficiency.

\section{Performance Across Target-Distractor Similarity}\label{app:distractor_similarity}
\begin{table}[t]
\centering\color{black}\small
\caption{Success rate for different levels of target and distractor similarity on CoIN-Bench Val Seen. \iclrrev{Results are obtained from separate evaluation runs using the same evaluation protocol as the main experiments.}}
\label{tab:rebuttal_similarity}
\begin{tabular}{lcccc}
\toprule
Similarity & Ours & AIUTA$^*$ & Pooled & $N$ \\
\midrule
LOW & \textbf{22.9} & 9.4 & 14.4 & 534 \\
MED & \textbf{18.5} & 10.3 & 13.0 & 146 \\
HIGH & \textbf{31.8} & 18.5 & 27.2 & 151 \\
\bottomrule
\end{tabular}
\end{table}

\iclrrev{We use Gemini 3 Flash Preview to group episodes by target-distractor similarity in color, texture, size, shape, and visible context. LOW, MED, and HIGH denote increasing similarity across these cues. ProCompNav outperforms both baselines across all three groups, including HIGH (Table~\ref{tab:rebuttal_similarity}).}

\section{\iclrrev{Human Evaluation Details}}\label{app:user_study}
\iclrrev{Both studies were conducted online using a within-subjects design. We pre-specified a target sample size of 20 participants per study and analyzed the first 20 completed sessions. The studies use target images and questions drawn from method logs. Questions were translated into Korean using Gemini~3 Flash Preview and manually checked before deployment.}

\noindent\textbf{\iclrrev{Recruitment and consent.}}
\iclrrev{Participants volunteered in response to recruitment announcements shared with colleagues and acquaintances. All were Korean nationals in their twenties or thirties and participated from South Korea. The samples may overlap across the two studies. Before starting, participants read a Korean-language explanation of the study purpose and procedure and provided electronic consent by clicking ``Agree and start.'' The specific hypotheses were not disclosed to reduce response bias. Incomplete sessions were excluded from the analysis.}

\noindent\textbf{\iclrrev{Duration and compensation.}}
\iclrrev{Recruitment announcements estimated approximately 10 minutes per study. Among the analyzed participants, the median elapsed time from consent to final completion was 7.5 minutes in Study~1 and 7.4 minutes in Study~2, including the tutorial, waiting, responses, and preference selection. No compensation was provided. In Study~2, participants were informed that waiting would occur and that they could do other work while waiting for a notification.}

\noindent\textbf{\iclrrev{Collected information.}}
\iclrrev{Study records contain randomly generated participant IDs, responses, response times, preference choices, and experiment metadata such as conditions, questions, and event timestamps. The exported records contain no names, email addresses, ages, genders, or IP addresses. Only aggregate participant results are reported in the paper. Appendix~\ref{app:user_study_instructions} provides English translations of the introductory instructions shown in the apps.}

\begin{figure}[H]
\centering
\includegraphics[width=\linewidth]{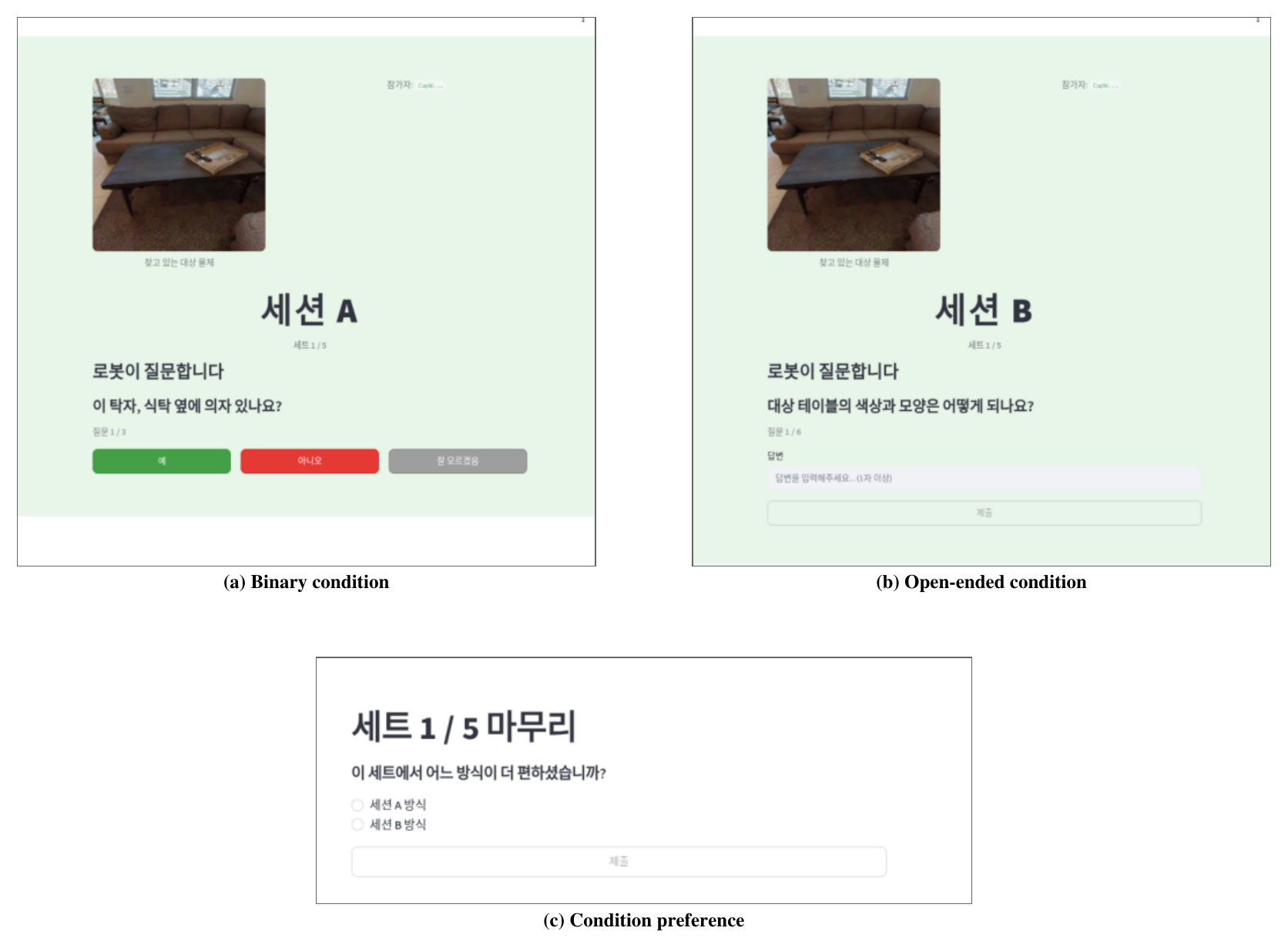}
\caption{Study~1 interface, translated into English and cropped, with text sizes adjusted to highlight the response controls. (a) The binary condition provides yes/no/unknown options. (b) The open-ended condition provides a free-text field. (c) Participants select their preferred condition after each paired trial. The study was conducted in Korean, and condition order was counterbalanced.}
\label{fig:user_study_ui}
\end{figure}

\noindent\textbf{\iclrrev{Study 1: question conditions.}}
\iclrrev{The binary-question condition presents yes/no/unknown buttons, whereas the open-ended-question condition presents a free-text field. Binary questions are drawn from ProCompNav logs, and open-ended questions from Pooled Independent Matching logs. Each participant completes five paired trials covering cabinet, bed, table, desk, and chair. All participants experience the same five target episodes, but the episode order varies across participants. Both conditions use batched delivery and the same target image within each pair. Whether binary or open-ended questions come first is balanced across participants to reduce order effects from familiarization or fatigue. Each participant answers 13 binary and 27 open-ended questions. We retain each method's original logged question sequences rather than equalizing question counts, preserving the question burden of each interaction strategy. Open-ended responses require at least one character, with no time, length, or keyword constraints. Figure~\ref{fig:user_study_ui} shows the interface used to answer both question types and report preference after each paired trial.}

\noindent\textbf{\iclrrev{Study 2: question delivery.}}
\iclrrev{Three binary questions are presented in each condition without requiring navigation. We sample waiting times independently and uniformly within ranges to approximate unpredictable question timing during robot interaction and reduce anticipation based on elapsed time. Participants see expected ranges rather than a countdown. The ranges are informed by Val Seen log medians, with exploration delays halved to limit study duration and LLM processing delays retained. The interleaved schedule is adjusted to keep total waiting time comparable across conditions.}

\iclrrev{The batched condition uses an initial wait of 157-173~s and between-question intervals of 4.75-5.25~s. The interleaved condition uses an initial wait and between-question intervals of 57-63~s. The two target images are balanced across the batched and interleaved conditions across participants. Condition order is counterbalanced, and participants select their preferred delivery after experiencing both.}

\noindent\textbf{\iclrrev{Measurement.}}
\iclrrev{Response time runs from question display to submission. Descriptive response times exclude outliers separately within each condition using the 1.5-IQR rule. Participant-level comparisons use each participant's median response time across all responses in each condition. In Study~1, every participant was faster in the binary condition. In Study~2, 12 of 20 participants were faster with batched delivery.}

\noindent\textbf{\iclrrev{Preference and response length.}}
\iclrrev{In Study~1, participant preference is the majority of five paired choices. The binary condition was preferred by all 20 participants and in 96 of 100 paired trials. In Study~2, 14 of 20 participants selected batched delivery after experiencing both conditions. Open-ended responses are segmented by whitespace: 481 of 540 (89.1\%) contain at most three words, with medians of one word and three characters. These are whitespace-delimited Korean words, not MLLM tokens, and are not directly comparable to the token-level Response Length reported for the user simulator.}

\subsection{\iclrrev{Participant Instructions}}\label{app:user_study_instructions}
\iclrrev{The following are English translations of the Korean introductory instructions shown before the ``Agree and start'' button. Icons and visual formatting are omitted. The recruitment announcements estimated 10 minutes for each study, while the Study~2 app estimated 8 minutes.}

\iclrrev{The accompanying code package includes the complete original Korean instructions, including tutorial and session-transition messages, in the study apps' \texttt{config.py} and \texttt{app.py} files under \texttt{user\_study/}.}

\begin{minipage}{\linewidth}
\noindent\textbf{\iclrrev{Study 1: introductory instructions.}}
\begin{quote}
\iclrrev{This experiment compares two ways in which a robot asks you questions. The two ways are presented in separate sessions, and you will complete five pairs of sessions.}

\iclrrev{Purpose: This experiment investigates which of two ways for a robot to ask users questions feels more psychologically comfortable.}

\iclrrev{Important notes: You may participate on either a PC or a mobile device. Please answer the robot's questions as conscientiously as possible. If you close the window, you cannot resume where you left off. Please complete the experiment in one sitting if possible.}

\iclrrev{Estimated time: Approximately 10 minutes.}

\iclrrev{Procedure: You will complete five sets. Each set concerns a different target object and consists of Session A, Session B, and a choice of which way was better. The target object image is always displayed at the top of the session. Depending on the session, you will answer by clicking a button (yes/no/don't know) or entering text. The experiment ends after all five sets.}

\iclrrev{Data: Only an anonymous ID and your responses are stored. Personally identifying information is not collected.}
\end{quote}
\end{minipage}

\noindent\textbf{\iclrrev{Study 2: introductory instructions.}}
\begin{quote}
\iclrrev{This experiment compares two ways in which a robot asks you questions. Each way is presented in a separate session.}

\iclrrev{Purpose: This experiment investigates which of two ways for a robot to ask users questions feels more psychologically comfortable.}

\iclrrev{Important notes: Please participate on a PC only. Mobile devices are not supported. Headphones or earphones are recommended so that you can hear notifications. You do not have to answer the robot's questions perfectly. Please answer comfortably, and incorrect answers are acceptable. If you close the window, you cannot resume where you left off. Please complete the experiment in one sitting if possible.}

\iclrrev{Waiting: You will spend some time waiting for the robot to explore within the page. Feel free to do other work while waiting. When the robot is ready to ask a question, a notification sound will play.}

\iclrrev{Estimated time: Approximately 8 minutes.}

\iclrrev{Procedure: You will complete two sessions. In each session, the robot briefly explores a room and then asks three yes/no questions about the target object. The robot asks questions to find the target object. You may do other work while it explores. Check the question when you hear the notification. The target object image is always displayed at the top of the session. Look at the image and answer each question with yes, no, or don't know. After the two sessions, complete a short survey about the two ways, and the experiment ends.}

\iclrrev{Data: Only an anonymous ID and your responses are stored. Personally identifying information is not collected. Your responses are not graded for correctness.}
\end{quote}

\noindent\textbf{\iclrrev{Response and preference instructions.}}
\iclrrev{In Study~1, the practice screens instruct participants to ``Answer by pressing one of the three buttons below'' or ``Enter an answer of at least one character in the field below, then submit.'' The main-trial instructions specify the sequence ``Session A: answer several questions about a target object. Session B: answer in a different way about the same target object. Choose which way was more comfortable.'' The preference question is ``Which way was more comfortable in this set?'' with options ``Session A way'' and ``Session B way.'' In Study~2, the tutorial explains that exploration in the main study takes approximately 57 seconds to 2 minutes 55 seconds depending on the session, but that the tutorial wait is shortened to 5 seconds. The preference screen states ``You have experienced both ways. Please choose one'' and asks ``Which of the two ways you just experienced do you prefer?'' Before choosing Session A or Session B, participants see each session described as either ``a long wait followed by several consecutive questions'' or ``a short wait followed by several questions at regular intervals.''}

\section{\iclrrev{Failure Analysis Details}}\label{app:failure_attribution}

\iclrrev{We examine where navigation fails and why RCJ may prune the target. Figure~\ref{fig:tom3_failure_modes} summarizes the failure stages, and Table~\ref{tab:rebuttal_pruning_failure} focuses on 134 target-pruning episodes. We use VLM judgments of candidate images to check whether the candidate finally selected as the target contains the intended instance. If it does but the episode fails, we count a navigation failure. These VLM checks are used only for analysis.}

\iclrrev{For each pruning episode, we give Gemini~3 Flash Preview\footref{fn:gemini_flash} the target image and ask whether it shows the attribute selected in the pruning round. In a separate call, we provide the target candidate's multi-view caption and ask whether it states or clearly implies the same attribute. We also inspect the target's group before and after refinement and the Qwen3-VL-8B response. Each episode is assigned to the first applicable category below.}

\begin{enumerate}
    \item \iclrrev{\textbf{Captions.} If Gemini judges that the caption describes the selected attribute but the target image does not show it, we assign \textbf{Caption Hallucination}. The reverse is \textbf{Caption Omission}. When both types occur in one episode ($n=1$), we count it under hallucination.}
    \item \iclrrev{\textbf{Core set.} If Gemini finds the selected attribute in neither the target image nor its caption, but the target is in the core set, we assign \textbf{Core-set False Positive}. A \textbf{No} answer from the user simulator then removes the core set, including the target.}
    \item \iclrrev{\textbf{Refinement.} For a target in the remainder set, we assign \textbf{Refinement False Negative} if Gemini finds the selected attribute in both the target image and its caption, but refinement does not move the target to the core set. We assign \textbf{Refinement False Positive} if Gemini finds it in neither, but refinement moves the target to the core set.}
    \item \iclrrev{\textbf{User simulator.} We assign \textbf{Simulator-Judge Mismatch} when Gemini's answer from the target image would retain the target, but the user-simulating MLLM Qwen3-VL-8B gives the opposite answer and the target is pruned.}
\end{enumerate}

\iclrrev{Although Gemini is a high-performing MLLM, it may also misread the target image or interpret the question differently from Qwen3-VL-8B. Their disagreement alone therefore does not mean Qwen3-VL-8B is wrong. When judging the target image, Gemini is instructed to answer \textbf{No} if the selected attribute is absent or cannot be assessed from that image. The multi-view caption uses the agent's collected images. An attribute may be present on the target but not visible from the captured viewpoints, leaving it absent from the caption.}

\par\medskip
\noindent\begin{minipage}{\linewidth}
\section{\iclrrev{Open-ended Feedback Corruption Example}}
\label{app:flip_examples}

\centering
\includegraphics[width=\linewidth]{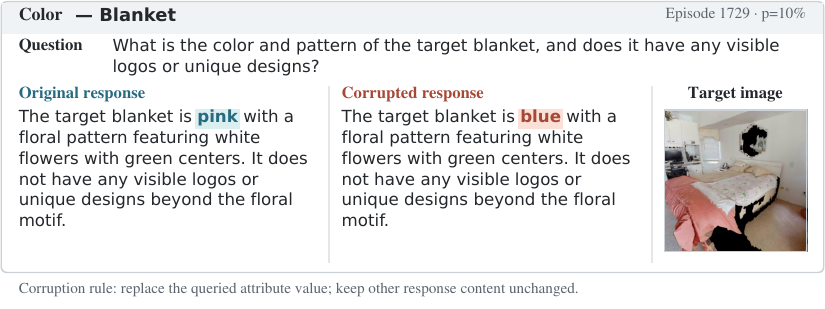}
\captionof{figure}{\iclrrev{An actual open-ended Flip example from AIUTA$^*$ Val Seen logs (episode~1729, $p=10\%$). The color value changes from pink (teal highlight) to blue (orange highlight), while all other response content remains unchanged. The target image is shown at right.}}
\label{fig:open_ended_flip_examples}
\end{minipage}
\par\medskip

\clearpage
\section{Dialogue Examples}\label{app:full_dialogue_comparisons}

\begin{figure}[H]
\centering
\includegraphics[width=\linewidth,height=0.9\textheight,keepaspectratio]{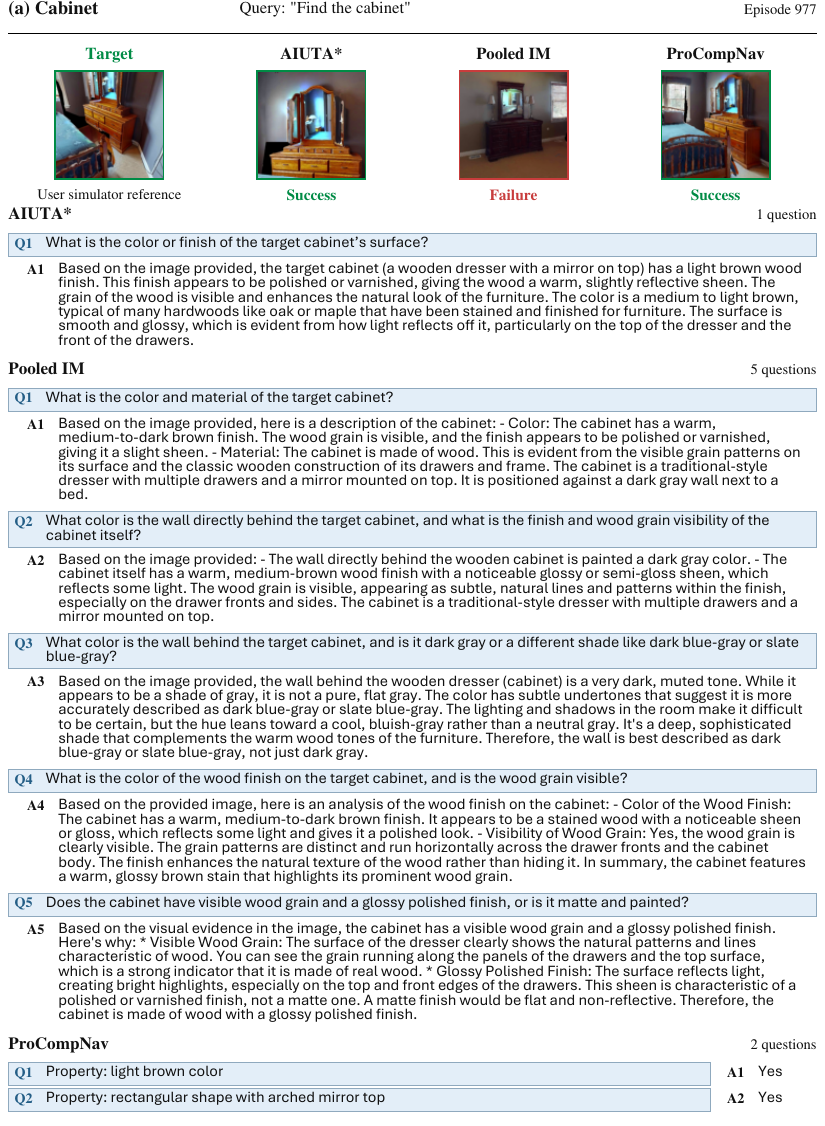}
\caption{Full dialogue comparison for episode 977. AIUTA* and ProCompNav succeed, whereas Pooled selects a distractor. All recorded user-simulator questions and answers are shown; ProCompNav questions are displayed as their recorded properties. The target image is the user simulator's reference. Pooled and ProCompNav views are reconstructed at logged poses; AIUTA* uses an approximate action replay.}
\label{fig:shared_attribute_examples}
\end{figure}

\begin{figure}[p]
\centering
\includegraphics[width=\linewidth,height=0.9\textheight,keepaspectratio]{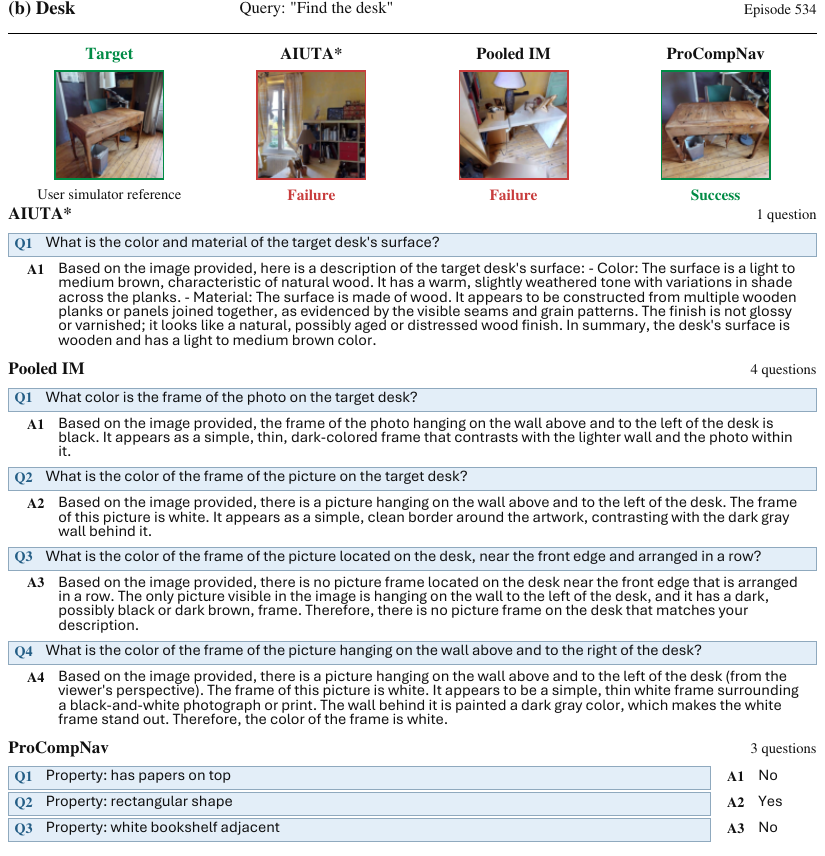}
\caption{Full dialogue comparison for episode 534. AIUTA* accepts a distractor after one question; Pooled asks four questions and selects an incorrect candidate, whereas ProCompNav succeeds with three binary questions. All recorded user-simulator questions and answers are shown; ProCompNav questions are displayed as their recorded properties. Pooled collects four candidates without the target; ProCompNav collects five candidates. The target image is the user simulator's reference. Method images are illustrative reconstructions from the episode-start pose (AIUTA*) and selected-group coordinates (Pooled and ProCompNav).}
\label{fig:shared_attribute_example_534}
\end{figure}
\clearpage

\section{\iclrrev{Asset Sources and Licenses}}\label{app:asset_licenses}

\iclrrev{Both navigation benchmarks use scenes from the Habitat-Matterport 3D Dataset (HM3D)~\citep{ramakrishnan2021hm3d}. Rendered target images are subject to the \href{https://matterport.com/legal/matterport-end-user-license-agreement-academic-use-model-data}{Matterport academic-use agreement}. Table~\ref{tab:asset_licenses} records the principal upstream sources and their publicly stated licenses or access terms.}

\begin{table}[H]
\centering
\small
\caption{\iclrrev{Principal asset sources and licenses. Asset names link to their original releases or license documents. A code license is distinguished from the terms for dataset annotations and scene data.}}
\label{tab:asset_licenses}
\begin{tabularx}{\linewidth}{@{}p{0.53\linewidth}X@{}}
\toprule
Asset or release & License or access terms \\
\midrule
\href{https://aihabitat.org/datasets/hm3d/}{HM3D v0.2 scenes} & Matterport academic-use agreement \\
\href{https://huggingface.co/datasets/ftaioli/CoIN-Bench}{CoIN-Bench v1 annotations} & Not specified in the dataset release \\
\href{https://github.com/XinyuSun/PSL-InstanceNav/blob/master/dataset.md}{PSL/TextNav annotations} & Not specified in the dataset release \\
\href{https://github.com/intelligolabs/CoIN/blob/master/LICENSE}{CoIN/AIUTA implementation} & CC BY-NC 4.0 \\
\href{https://github.com/rai-opensource/vlfm/blob/main/LICENSE}{VLFM implementation} & MIT \\
\href{https://github.com/facebookresearch/habitat-sim/blob/main/LICENSE}{Habitat-Sim 0.2.4} & MIT \\
\href{https://huggingface.co/Qwen/Qwen3-VL-8B-Instruct-GGUF}{Qwen3-VL-8B-Instruct} & Apache 2.0 \\
\href{https://huggingface.co/MoritzLaurer/DeBERTa-v3-large-mnli-fever-anli-ling-wanli}{DeBERTa-v3-large} & MIT \\
\href{https://huggingface.co/facebook/dinov2-large}{DINOv2-large} & Apache 2.0 \\
\href{https://huggingface.co/sentence-transformers/all-MiniLM-L6-v2}{all-MiniLM-L6-v2} & Apache 2.0 \\
\href{https://github.com/IDEA-Research/GroundingDINO/blob/main/LICENSE}{GroundingDINO Swin-T} & Apache 2.0 \\
\href{https://github.com/ChaoningZhang/MobileSAM/blob/master/LICENSE}{MobileSAM} & Apache 2.0 \\
\href{https://huggingface.co/Salesforce/blip2-itm-vit-g}{BLIP2 ITM ViT-g} & MIT \\
\href{https://ai.google.dev/gemma/terms}{Gemma-3-12B} & Gemma Terms of Use \\
\href{https://blog.google/products-and-platforms/products/gemini/gemini-3-flash/}{Gemini 3 Flash Preview} & Hosted API, subject to provider terms \\
\href{https://help.aliyun.com/en/model-studio/qwen3-6-plus}{Qwen3.6-Plus} & Hosted API, subject to provider terms \\
\bottomrule
\end{tabularx}
\end{table}

\section{Prompts}\label{app:prompts}
We provide the prompt templates used in ProCompNav and Pooled Independent Matching. Red placeholders are replaced with the corresponding inputs.

\subsection{ProCompNav Prompts}

The multi-view description prompt is shared by CoIN-Bench and TextNav; the remaining prompts are grouped by setting below.

\noindent\textbf{Multi-view Object Description:}
\par\nopagebreak[4]

\promptbox{
You are given an image which shows a \promptvar{category} from multiple viewpoints. You need to describe the \promptvar{category}, by combining the information from these different viewpoints.\\
First, describe the \promptvar{category} in detail, focusing on its appearance and distinctive features (use only: color, shape).\\
Then, describe the other objects (use only: color, shape) close to the \promptvar{category}, and their spatial relationships (use only: [`next to', `on top of', `under']) relative to the \promptvar{category}.\\
Mention all clearly visible nearby objects around the \promptvar{category}.
}

\vspace{0.4em}
\noindent
This prompt is used to generate a unified caption from multiple viewpoints of the same object instance.

\subsubsection{CoIN-Bench}

\noindent\textbf{Shared Attribute Extraction:}
\par\nopagebreak[4]

\promptbox{
You are given a list of [\promptvar{category}] descriptions.\\
Task: Identify properties that are shared by ALL [\promptvar{category}s] in the list.\\
- Properties should be based on visible appearance or nearby surrounding objects.\\
- If a surrounding object is distinctive, explicitly name it.\\
- Each property should be specific but concise.\\
- Each property must be no more than 10 words.\\
- Output only properties, one per line.\\
\\
List of descriptions:\\
\promptvar{descriptions}\\
\\
DO NOT USE the following properties: [\promptvar{used\_properties}].\\
Each property must apply to every object in the list.\\
Extract at most 10 valid shared properties.\\
\\
Format:\\
property1\\
property2\\
property3
}

\vspace{0.4em}
\noindent
where \promptvar{descriptions} is replaced by the candidate descriptions and
\promptvar{category} by the target instance category $c$. The used-property list excludes attributes already queried.

\vspace{1em}









\noindent\textbf{Binary User Simulation:}
\par\nopagebreak[4]

\promptbox{
image: <image>\\
Does this [\promptvar{category}] have the following property?\\
PROPERTY: [\promptvar{property}]\\
Answer strictly with `yes' or `no'. Do not say anything else.
}

\vspace{0.4em}
\noindent
The image is the target-instance image available only to the user simulator. The selected attribute-value pair replaces \promptvar{property}.

\subsubsection{Text-goal Navigation (TextNav)}

\noindent\textbf{Text-goal Attribute Extraction:}
\par\nopagebreak[4]

\promptbox{
You are given a single [\promptvar{category}] description.\\
Task: Extract distinctive properties of this [\promptvar{category}].\\
- Properties must be specific to this [\promptvar{category}], not generic.\\
- Include only directly observable physical attributes of the [\promptvar{category}] itself or objects adjacent to it.\\
- Exclude atmosphere, emotional tone, interpretive statements, and descriptive effects.\\
- Each property should be specific but concise.\\
- Each property must be no more than 10 words.\\
- Output only properties, one per line.\\
\\
Description:\\
\promptvar{text\_goal}\\
\\
Extract at most 10 valid properties.\\
\\
Format:\\
property1\\
property2\\
property3
}

\vspace{0.4em}
\noindent
This prompt extracts the fixed attribute set from the initial TextNav goal description.

\vspace{1em}

\noindent\textbf{Text-goal Matching:}

\promptbox{
You are given two descriptions of a [\promptvar{category}].\\
Description A (reference): \promptvar{reference\_description}\\
Description B (candidate): \promptvar{candidate\_description}\\
Notes:\\
- Each description can include both the object description and nearby surrounding context.\\
- Focus on whether they refer to the same object; surrounding context may differ.\\
Question: Do Description A and Description B refer to the same [\promptvar{category}]?\\
Answer strictly with `yes' or `no'. Do not say anything else.
}

\vspace{0.4em}
\noindent
This prompt is used to check whether a final candidate matches the given detailed description in the TextNav setting.

\subsection{Pooled Independent Matching Prompts}
These prompts are used for Pooled Independent Matching on CoIN-Bench.

\noindent\textbf{Target Fact Update:}
\par\nopagebreak[4]

\promptbox{
You are an intelligent embodied agent tasked with finding a specific target \promptvar{category}.\\
\\
You know the following facts about the target \promptvar{category}:\\
<START\_TARGET\_PICTURE\_FACTS> \promptvar{target\_facts} <END\_TARGET\_PICTURE\_FACTS>\\
\\
You recently detected another picture and asked to the human several question:\\
<START\_OF\_ORACLE\_ANSWER>\\
\promptvar{question\_answer\_pairs}\\
<END\_OF\_ORACLE\_ANSWER>\\
\\
Task: Update the target facts with this new information. Be concise. Do not include information that are uncertain.\\
\\
YAML\_START\\
facts: <updated facts as a single text line>\\
YAML\_END \# must be present to get the information back\\
Provide your reasoning step-by-step, after the YAML\_END tag.
}

\vspace{0.4em}
\noindent
This AIUTA prompt updates the accumulated facts after each user response. Each question-answer pair is inserted as \texttt{- Question: <question> - Answer: <answer>}.

\vspace{1em}

\noindent\textbf{Tie-breaking Candidate Selection:}
\par\nopagebreak[4]

\promptbox{
collected target fact: \promptvar{target\_facts}\\
\\
\promptvar{candidate\_list}\\
\\
Pick the ONE candidate whose description best matches the target fact.\\
Respond with only the id. No other text.
}

\vspace{0.4em}
\noindent
When multiple candidates share the highest final score, \promptvar{candidate\_list} contains their descriptions, each prefixed with \texttt{candidate <id>:}, and the LLM selects one candidate.

\vspace{1em}

\noindent\begin{minipage}{\linewidth}
\noindent\textbf{Candidate Scoring and Question Generation:}
This single prompt, inherited from AIUTA~\citep{taioli2025collaborative}, returns both a matching score and a question. During the first pass, its question is used to collect user feedback. During the second pass, its score is used to rank candidates against the final facts; the generated question is not asked.

\promptbox{
You are an intelligent agent equipped with an RGB sensor, object detector, and Visual Question Answering (VQA) model.\\
Your goal is to identify a target \promptvar{category} based on a scene description and prior knowledge of the target.\\
\\
Scenario:\\
The object detector has identified a scene containing a \promptvar{category}, and the VQA model has provided the following description:\\
\\
<START\_OF\_DESCRIPTION>\\
\promptvar{candidate\_description}\\
<END\_OF\_DESCRIPTION>\\
\\
Target object information:\\
Based on previous interactions, you know the target picture has the following characteristics:\\
<START\_TARGET\_PICTURE\_FACTS>\\
\promptvar{target\_facts}\\
<END\_TARGET\_PICTURE\_FACTS>\\
\\
Task:\\
1. Similarity analysis.\\
Analyze how closely the detected scene description aligns with the known facts about the target \promptvar{category}. Provide a similarity score between 0 and 10, where:\\
- 0 = The detected \promptvar{category} is not the target object.\\
- 10 = The detected \promptvar{category} is definitely the target object.\\
- If no information about the target is available, the score should be -1.\\
\\
2. Question Generation:\\
- The question is for the target object, not the detected one.\\
- Ask exactly one specific, relevant, and human-answerable question related to the target object that maximizes information gain for identifying the target \promptvar{category}.\\
- Do not ask speculative or irrelevant questions\\
- The question should be grounded in observable or known details from the scene, focusing on key characteristics that can help confirm or refute the identity of the target object.\\
\\
Ensure your response follows the format below:\\
YAML\_START \# must be present to get the information back\\
similarity\_score: <similarity score>\\
questions:\\
\hspace*{1em}<question\_number>: <question\_content>\\
YAML\_END \# must be present to get the information back\\
\\
Provide your reasoning step-by-step for the similarity score and questions, after the YAML\_END tag.
}
\end{minipage}

\vspace{1em}

\noindent\textbf{Open-ended User Simulation:}
Following AIUTA~\citep{taioli2025collaborative}, the generated question itself is passed to the MLLM together with the target-instance image, without an additional response instruction. We use Qwen3-VL-8B as described in Section~\ref{sec:exp_setup}.


\end{document}